\documentclass{article}

\PassOptionsToPackage{numbers, compress}{natbib}
\usepackage[preprint]{neurips_2025}

\usepackage[utf8]{inputenc} 
\usepackage[T1]{fontenc}    
\usepackage{hyperref}       
\usepackage{url}            
\usepackage{booktabs}       
\usepackage{amsfonts}       
\usepackage{nicefrac}       
\usepackage{microtype}      
\usepackage[table,dvipsnames]{xcolor}    
\usepackage{amsmath}
\usepackage{graphicx}   
\usepackage{caption}    

\usepackage{enumitem}

\definecolor{mygreen}{rgb}{0.0, 0.5, 0.0} 

\newcommand{\gcheck}{\textcolor{mygreen}{\normalsize\checkmark}}

\usepackage{xspace}                
\newcommand{\TotalData}{1,211\xspace}

\RequirePackage{caption}
\DeclareCaptionLabelSeparator{custom}{}
\DeclareCaptionFormat{custom}{{\sffamily\textbf{#1 #2}} #3}
\usepackage{todonotes}
\usepackage{multirow}
\usepackage{wrapfig}
\renewcommand\paragraph[1]{\vspace{0em}\noindent\textbf{#1}}
\usepackage{float}

\usepackage{tabularx}

\usepackage{tabularx}
\usepackage{caption}
\usepackage{booktabs}

\newcolumntype{L}{>{\raggedright\arraybackslash}X}

\title{Seeing is Not Reasoning: MVPBench for Graph-based Evaluation of Multi-path Visual Physical CoT}

\newcommand{\benchmark}{MVPBench}

\author{
Zhuobai Dong $^1$$^*$,~
Junchao Yi $^2$$^*$,~Ziyuan Zheng $^1$,~
Haochen Han$^3$,~
Xiangxi Zheng $^4$,
\\
\textbf{Alex Jinpeng Wang $^1$$^\dagger$,~
Fangming Liu $^3$,~
Linjie Li $^5$~}
\\
$^1$ Central South University
$^2$ University of Electronic Science and Technology of China
\\
$^3$ Peng Cheng Laboratory
$^4$ Nanjing University
$^5$ Microsoft
}

\begin{document}
\maketitle
\vspace{-2em}
\begin{center}
\texttt{Homepage: \url{https://csu-jpg.github.io/MVPBench/}}
\end{center}

\let\thefootnote\relax\footnotetext{$^*$Equal contribution, $^\dagger$Corresponding author}

\begin{abstract}
Understanding the physical world—governed by laws of motion, spatial relations, and causality—poses a fundamental challenge for multimodal large language models (MLLMs).
While recent advances such as OpenAI o3 and GPT-4o demonstrate impressive perceptual and reasoning capabilities, our investigation reveals these models struggle profoundly with visual physical reasoning, failing to grasp basic physical laws, spatial interactions, and causal effects in complex scenes. More importantly, they often fail to follow coherent reasoning chains grounded in visual evidence, especially when multiple steps are needed to arrive at the correct answer.
To rigorously evaluate this capability, we introduce \textbf{MVPBench}, a curated benchmark designed to rigorously evaluate visual physical reasoning through the lens of visual chain-of-thought (CoT). 
Each example features interleaved multi-image inputs and demands not only the correct final answer but also a coherent, step-by-step reasoning path grounded in evolving visual cues. 
This setup mirrors how humans reason through real-world physical processes over time.
To ensure fine-grained evaluation, we introduce a \textbf{graph-based CoT consistency metric} that verifies whether the reasoning path of model adheres to valid physical logic.
Additionally, we minimize shortcut exploitation from text priors, encouraging models to rely on visual understanding.
Experimental results reveal a concerning trend: even cutting-edge MLLMs exhibit poor visual reasoning accuracy and weak image-text alignment in physical domains. 
Surprisingly, \textbf{RL-based post-training alignment—commonly believed to improve visual reasoning performance—often harms spatial reasoning}, suggesting a need to rethink current fine-tuning practices.
\end{abstract}

\section{Introduction}
\label{intro}
Human comprehension of the world is fundamentally grounded in physical laws: objects fall when released, and liquids take the shape of their containers~\cite{spelke1992origins, baillargeon2004infants}.
Such physical regularities form the basis of our causal understanding~\cite{gopnik2004theory, lake2017building}, and further link the chain of reasoning when solving complex problems.
Recent advances appear to grasp this physical world that humans experience—a blitz of multimodal large language models (MLLMs) like OpenAI o3~\cite{openai2025gpto3},
GPT4o~\cite{openai2024gpt4o}, Gemini~\cite{geminiteam2024geminifamilyhighlycapable}, InternVL3~\cite{InternVL3}, Kimi1.5~\cite{kimiteam2025kimik15scalingreinforcement} and many others\cite{chen2025r1v,zheng2025easyr1}
-all claiming \emph{human-level physical reasoning} after a final reinforcement-learning (RL) post-training.
Recent works \cite{shao2024visual,guo2025generateimagescotlets,li2025imaginereasoningspacemultimodal,daxberger2025MM-Spatial,huang2025MLLM-For3D,fan2024MLLM-SUL} show models describing panoramic scenes, solving game reasoning, even generating Chain-of-Thought (CoT) explanations.
At first glance, it feels as thought \textbf{plug-and-play embodied intelligence is already on our doorstep.}

\begin{figure*}[t]
    \centering
    \includegraphics[width=\linewidth]{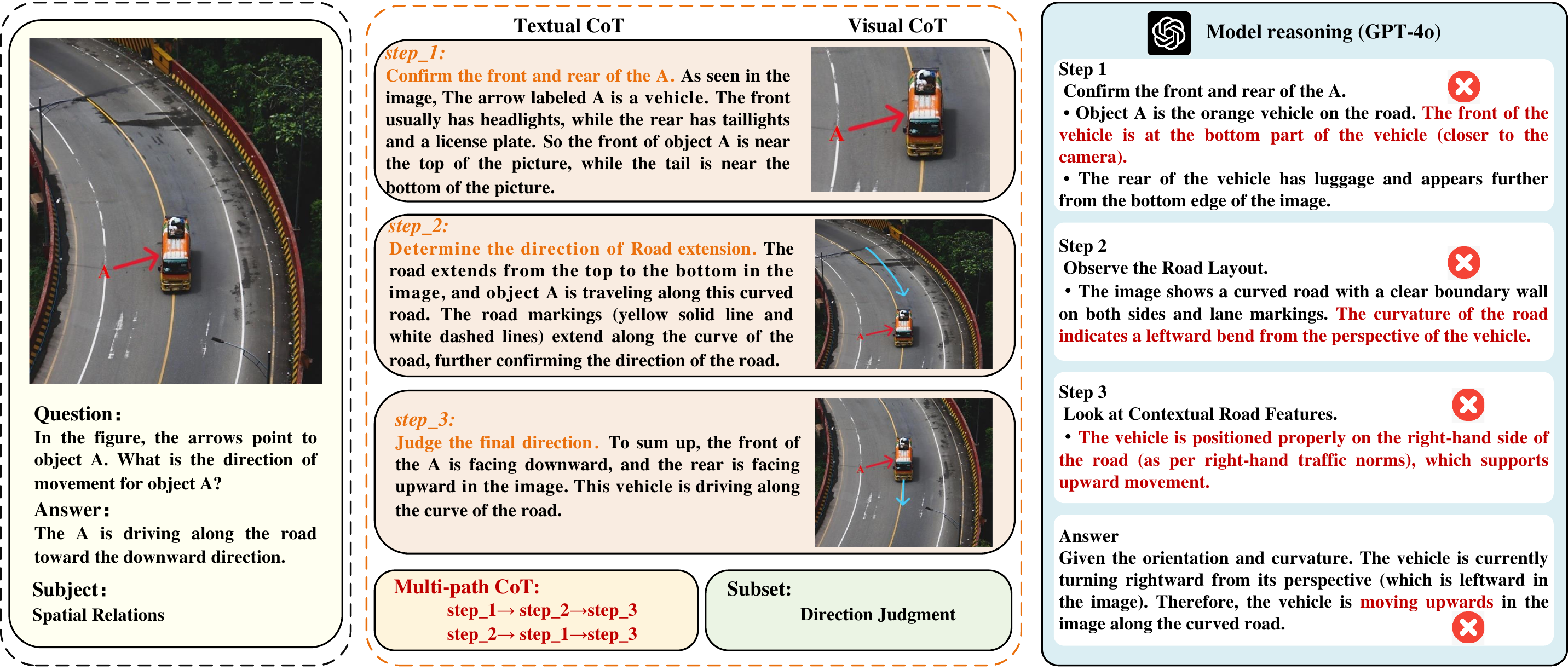}
    \caption{
    \textbf{A one-minute sanity check shatters the illusion \emph{of spatial reasoning in MLLMs}.}
    Red arrows indicate objects and multiple reasoning chains are provided to capture diverse yet valid solution strategies.
    }
    \label{fig:motivation1}
\end{figure*}

Full of eager expectation, we asked the lastest MLLMs a child-level physics question.
\emph{What is the direction of movement for the car?}
Fig.~\ref{fig:motivation1}(left) shows the setup.
Surprisingly, GPT-4o responded with an incorrect prediction.
Pushing further, we queried the thought chain of models.
The failure patten was consistent: models \emph{saw} the pixels but did not \emph{reason} about forces, geometry, or causality.

“The Second Half,” reminds us AI is entering a phase where evaluation outweighs training\cite{yao2025secondhalf}. 
Yet current benchmarks used to “prove” spatial reasoning are a weak compass.  
Most rely on game-engine videos or CAD renderings whose textures and lighting barely resemble the messy real world\cite{kang2024how,zheng2024contphy}.  
In addition, many questions are phrased so that a language-only model can guess the answer from commonsense priors, bypassing vision altogether\cite{zhou2025MDK12-Bench,yue2024MMMU,lee2025vlindbenchmeasuringlanguagepriors}.  
Furthermore— almost none pair each intermediate visual cue with an explicit reasoning step, so training pipelines receive no pressure to ground chain-of-thought in what the model \emph{sees};\cite{jiang2025mme,zhang2024mathverse,shao2024visual} RL post-training therefore optimizes conversational fluency while silently tolerating physical implausibility.  
The result is \textbf{a generation of MLLMs that can \emph{describe} images eloquently yet still misjudge which way a car is moving}.

To close this evaluation gap, we introduce \textbf{MVPBench}, a \underline{M}ulti-path \underline{V}isual \underline{P}hysics benchmark that turns the spotlight on vision-centric reasoning. 
MVPBench contains \TotalData carefully curated examples across three real-world domains:  
\emph{i.} hands-on physics experiments (electromagnetic induction, heat conduction, collisions),
\emph{ii.} exam-style word problems requiring symbolic or commonsense reasoning, and  
\emph{iii.} spatial-transformation tasks that challenge 3D understanding (viewpoint shifts, object rearrangement).  
Each example pairs \emph{multi-image evidence} with \emph{multiple valid CoT paths}, forcing models to justify every step in view of changing visuals.
To evaluate such rich annotations, we introduce a \textbf{graph-based CoT metric suite} that represents each reasoning chain as a directed acyclic graph of atomic facts and then assesses step-wise fidelity through exact or fuzzy graph matching, measures text–image grounding with automated alignment scores, and quantifies multi-path coverage by rewarding diverse yet logically valid reasoning flows.
MVPBench thus re-aligns the compass: genuine physical understanding demands that models \emph{see, think, and prove}—not merely narrate.

Extensive experiments reveal two key insights:
\emph{i}. Providing models with the full image sequence boosts performance by up to 21\% points-evidence that temporal context matters.
\emph{ii}. Contrary to conventional wisdom, RL-based post-training \emph{reduces} visual-physics scores on MVPBench by 2\% points, indicating that current reward designs sacrifice grounded reasoning for coherence.

\textbf{To summarize, this paper makes the following contributions:}\emph{i}. To the best of our knowledge, \textbf{MVPBench} is the first benchmark to combine real-world visual physics, multi-image inputs, and \emph{multi-path} CoT annotations. \emph{ii}. A \textbf{graph-based evaluation toolkit} that jointly measures reasoning fidelity, visual grounding, and path diversity.
\emph{iii}. The first comprehensive study showing that widely adopted RL alignment can impair spatial reasoning, calling for vision-centric reward design.

\section{Related Works}
\label{related_works}
\paragraph{Limitations of Multi-modal Large Language Models.} Although MLLMs have made significant progress, recent studies have found that their understanding of the physical world is still weak\cite{MOKA,PhyGrasp,evaluatingmultiviewobjectconsistency}, and they face major challenges in reasoning based on visual perception\cite{openeyesreason,V-MAGE,VERIFY}.
In terms of physical discipline knowledge, the ability of model to perform multimodal reasoning is limited\cite{OlympiadBench}.
When faced with tasks involving the prediction of physical interactions, the model shows insufficient understanding\cite{CLEVRER,Physion}. 
Additionally, there are obvious deficiencies in the model to accurately interpret object properties and states in physics-based scene evaluations\cite{NEWTON,physicscontextbuildersmodular}.
Although the spatial reasoning ability of mllm is constantly improving, it still often struggles to understand spatial relationship problems through visual perception and reasoning\cite{SpatialVLM}. 
These findings emphasize the need for more comprehensive and rigorous benchmarks specifically designed to evaluate visual reasoning capabilities of mllms in physical understanding.

\paragraph{Physical Comprehension Datasets.} These datasets have become a crucial area of focus, posing a significant challenge for MLLMs. 
Early physical benchmarks\cite{Physion,benchmarkingsequentialvisualinput,Physion++}were developed around simple physical scene reasoning. 
Inspired by research on infant intuitive physics, the study\cite{IntPhys} evaluate innate understanding of models in the physical world. 
In other aspects of physical datasets, existing benchmarks\cite{OlympiadBench,VisScience,lu2022learnexplainmultimodalreasoning,EMMA,PhysReason} to evaluate physics problems mainly focus on commonsense reasoning based on language knowledge. 
Spatial benchmarks\cite{EmbodiedScan,yang2024thinkingspacemultimodallarge,shiri2024empiricalanalysisspatialreasoning,li2024proximityqaunleashingpower}, on the other hand, emphasize spatial perception and reasoning in 3D scenes, illustrating the early stages of world model. 
Recent effort\cite{PhysBench} has expanded to comprehensively assess  understanding of models in physical scenes across various tasks, though they still fail to fully encompass real-world physical knowledge.
\textbf{By introducing visual CoT as inputs in a vision-centric manner, it forces models to reason across images, making it a closer approximation to the analysis of complex physical scenes in the real world.}

\paragraph{Multi-modal Large Language Models.} MLLMs extend traditional LLMs\cite{gpt4technicalreport,mixtralexperts} and vision models\cite{kirillov2023segany,zhang2023personalizesegmentmodelshot} to address diverse tasks across various modalities, including 2D images\cite{Flamingo,InstructBLIP}, 3D point clouds\cite{3D-LLM}, audio\cite{PandaGPT,fu2024vita,fu2025vita,VITA-Audio}, and videos\cite{Video-LLaMA,SpaceVLLM}. 
Notable models such as GPT-4o\cite{openai2024gpt4o} and Gemini\cite{geminiteam2024geminifamilyhighlycapable} have demonstrated exceptional visual reasoning capabilities, establishing new benchmarks in the field, though their closed-source nature limits broader accessibility and application. 
In contrast, open-source initiatives like LLaMA-Adapter\cite{zhang2024llamaadapter}, LLaVA\cite{liu2023llava,liu2023improvedllava,liu2024llavanext}, and MiniGPT-4\cite{zhu2023minigpt,chen2023minigptv2},integrate vision models like CLIP\cite{CLIP} for multimodal fine-tuning. 
While more recent models such as mPLUG-Owl\cite{mPLUG-Owl3}, Qwen2.5-VL\cite{Qwen2.5-VL}, InternVL3\cite{InternVL3} and many other works\cite{wu2024deepseekvl2,kimiteam2025kimivltechnicalreport,guo2025seed15vltechnicalreport,chris2025skyworkr1v2} are pushing the boundaries of MLLMs in visual understanding.
These advancements highlight the growing potential of MLLMs. 
However, their capabilities in visual physical reasoning remain underexplored. 
\textbf{In this paper, we introduce the \benchmark\ benchmark and provide a unique evaluation suite to comprehensively assess the visual reasoning capabilities of MLLMs in understanding physical world, thereby offering a distinctive perspective for guiding future research.}

\begin{table}[t]
\caption{
\textbf{Comparison of MVPBench with existing benchmarks for physical understanding.}
MVPBench covers a broader range of physical reasoning categories, supports multi-perspective chain-of-thought evaluation, and provides CoT annotations.
In the data format, TC indicates that the dataset utilizes textual CoT, VC means the use of visual CoT as input, and Vc signifies all that the data is constructed in a vision-centric manner. 
}
\vspace{\baselineskip}
\label{sample-table}
\centering
\scriptsize
\setlength{\tabcolsep}{2.5pt} 
\renewcommand{\arraystretch}{1.15} 
\resizebox{\textwidth}{!}{
\begin{tabular}{l|cccc|ccc|ccc}
  \toprule
  \multirow{2}{*}{\textbf{Benchmark}} & \multicolumn{4}{c|}{\textbf{Data category}} & \multicolumn{3}{c|}{\textbf{CoT Evaluation}} & \multicolumn{3}{c}{\textbf{Data format}} \\
   & Physics experiments & Physics problems & Spatial relations & Dynamic prediction & Quality & Diversity & Efficiency &  Vc & TC & VC\\
   
  \midrule
  PhysBench\cite{PhysBench}  &  & &  \gcheck & \gcheck & & & & \gcheck & & \\
  Physion\cite{Physion}    &  & & &  \gcheck & & & & & &\\
  PhysReason\cite{PhysReason} &  &  \gcheck & & & \gcheck &  &  &  & \gcheck & \\
  PhysGame\cite{cao2024physgame}   &  & & & \gcheck & & & &\gcheck & & \\
  ContPhy\cite{zheng2024contphy}    &  & & &  \gcheck &   & & & & & \\
  EmbSpatial\cite{du2024embspatial} &  & &  \gcheck & &  & & & & & \\
  \midrule
  \textbf{MVPBench}   & \gcheck & \gcheck & \gcheck & \gcheck & \gcheck & \gcheck & \gcheck & \gcheck & \gcheck & \gcheck\\
  \bottomrule
\end{tabular}
}


\end{table}

\section{MVPBench}
\label{headings}
The motivation behind constructing the \benchmark\ benchmark stems from recognizing significant gaps in the current capability of MLLM to deeply comprehend and reason about the physical world. 
Existing benchmarks predominantly emphasize isolated aspects such as static scene understanding, physics-based reasoning, or basic spatial awareness, leaving unaddressed the comprehensive integration of physical reasoning with complex visual inputs.
Therefore, \benchmark\ aims to rigorously evaluate abilities of MLLMs to visually reason about diverse physical phenomena in scenarios closely resembling real-world complexities.

To ensure comprehensive coverage of visual reasoning skills, \benchmark\ incorporates carefully curated data across multiple distinct yet complementary domains: 1) \emph{Physics Experiments} tests the understanding of sequential physical processes through multi-step visual inference. 
2) \emph{Physics Problems} challenges models to interpret advanced, visually grounded physics questions from academic examinations. 
3) \emph{Spatial Relations} assesses spatial perception judgment across various scenarios. 
4) \emph{Dynamic Prediction} evaluates the predictive capabilities of models regarding dynamically evolving physical interactions. 
Collectively, these diverse yet targeted subdomains ensure MVPBench not only addresses existing evaluation gaps but also significantly extends the reasoning depth, robustness, and versatility of models.
Details of data analysis are provided in Appendix C.

\begin{figure*}[t]
    \centering
    \includegraphics[width=\linewidth]{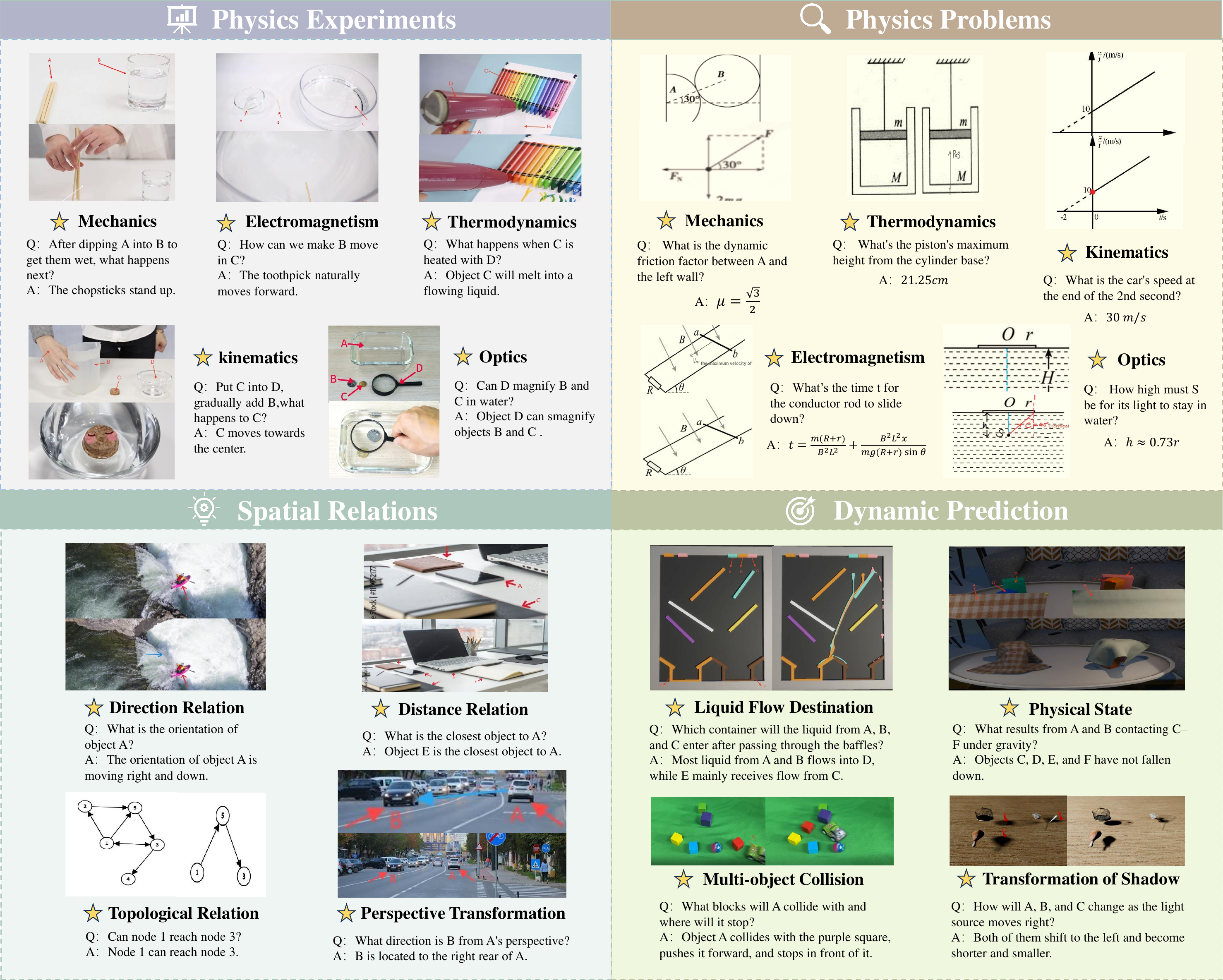}
    \caption{
    \textbf{Examples from MVPBench across four major categories.}
    Each example includes an initial scene followed by multiple reasoning steps. Target objects are marked with red arrows and labeled with letters to reduce textual bias.
    }
    \label{fig:all_datasets_visualization}
\end{figure*}

\subsection{Data Generation}
 
\paragraph{Physics Experiments.}
We scraped publicly available physics experiment videos, manually filtered them, and archived the curated clips as MP4 files.
From each video, we extracted key frames depicting (i) the initial setup, (ii) critical intermediate steps, and (iii) the final results. Salient objects were highlighted with arrows while all textual cues were omitted, forcing models to infer solely from visual cues, with GPT-4 generating the corresponding scene descriptions.
The intermediate steps encompass essential logical reasoning processes required to complete each experiment. 
\textbf{\emph{To evaluate multi-path reasoning verification capability of MLLMs, we recorded multiple chains of thought for each instance.}}
All assets are stored in a structured JSON schema that includes mechanics, thermodynamics, electromagnetism, optics and kinematics.
The remaining subsets employ the same JSON schema as the one detailed above and likewise store multiple reasoning paths.
Therefore, we omit related discussion in the following sections.

\paragraph{Physics Problems.}
On one hand, we crawled and manually filtered all the problems from relevant websites, compiling them into PDF files, which were then converted into Markdown format via OCR and manually aligned.
On the other hand, the data was  with examples from the PhysReason-mini\cite{PhysReason} dataset.
All problems are tightly coupled to images and drawn from examinations in several countries (predominantly Chinese college entrance examination) for their open-ended formats that demand advanced reasoning.
After meticulous verification, we extracted key reasoning steps and final answers.
These steps include both textual and visual components, with the image segment forming an additional input alongside the original image. 
The questions cover five subcategories including mechanics, thermodynamics, electromagnetism, optics, and kinematics.

\paragraph{Spatial Relations.}
Spatial relation reasoning is a crucial area in understanding of the physical world.
To address this gap, we have pre-designed four main subcategories to evaluate  perception of spatial relations: 
\textbf{(1) Direction judgment}: This subcategory formulates problems concerning the directional judgment of various objects.
\textbf{(2) Distance estimation}: This subset encompasses problems related to estimating the distance relation of different objects.
\textbf{(3) First view transformation}: This subcategory addresses issues pertaining to direction judgment from a egocentric viewpoint regarding various objects.
\textbf{(4) Topological relation judgment}: This subcategory focuses on problems associated with reachability within directed graphs.
The first three subcategories manually screened original images from public websites, and the fourth subcategory constructed images using the Graph Editor tool.

\paragraph{Dynamic Prediction.}
To investigate whether MLLMs can predict time-varying physical outcomes through visual reasoning, we introduce a Dynamic Prediction subset comprising four subcategories: \textbf{Multi-object Collision, Liquid Diversion, Physical state and Shadow Transformation} predict.
This subset utilizes the PhysBench \cite{PhysBench} benchmark, which provides high-quality dynamic scene videos. 
All samples are adapted and extended from PhysBench to ensure high-quality video frames.
For each sample, we extract multiple temporally spaced key frames from the corresponding video to form multi-image inputs, annotating salient objects with arrows.

\section{CoT Evaluation Method}
\label{others}

\begin{figure*}[t]
    \centering
    \includegraphics[width=\linewidth]{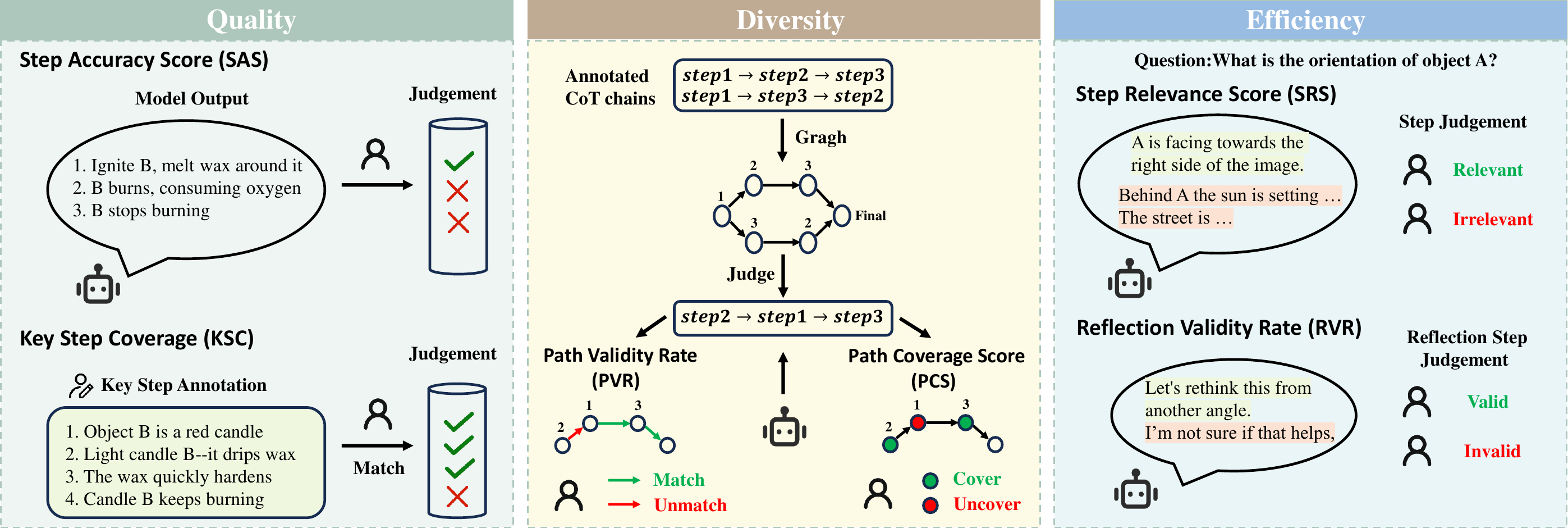}
    \caption{
    \textbf{Evaluation framework for multi-path Chain-of-Thought (CoT) reasoning.}
    MVPBench introduces a comprehensive protocol to evaluate CoT reasoning from three perspectives: quality, diversity, and efficiency.
    \textbf{For CoT diversity, we propose a graph-based multi-path evaluation method} that quantifies the ability of a model to explore alternative reasoning routes via Path Validity Rate (PVR) and Path Coverage Score (PCS), advancing beyond prior single-path metrics.
    }
    \label{fig:evaluation_framework}
\end{figure*}

Existing CoT evaluation methods often simplify reasoning assessment to a binary judgment of the final answer, overlooking the internal reasoning steps. 
To address this limitation, we propose a holistic CoT evaluation suite that captures the reasoning process across multiple dimensions, offering a finer-grained understanding of reasoning capabilities of MLLMs. 
\textbf{\textit{Notably, we are the first to introduce an evaluation metric for assessing multi-path reasoning ability of models, which complements traditional correctness and reflection assessments.}}
Details are presented in Section~\ref{subsec:correct} (correctness), Section~\ref{subsec:multisolu} (multi-path reasoning), and Section~\ref{subsec:reflection} (reflection quality).

\subsection{CoT Quality Evaluation}
\label{subsec:correct}

To evaluate the correctness of CoT reasoning, we extend existing interpretable metrics by incorporating both step-wise accuracy and final answer correctness. 
While prior work such as \cite{jiang2025mme} focused on intermediate informativeness, they overlook the contribution of the final answer to overall quality. 
Inspired by \cite{zhang2024mathverse}, we introduce a \textbf{weighted scoring framework} that balances the quality of intermediate steps with the correctness of the final prediction.

\paragraph{Step Accuracy Score (SAS).}
We prompt GPT-4o~\cite{openai2024gpt4o} to decompose each CoT prediction into steps, categorized as logical inference, image captioning, or background/numerical computation (depending on the dataset). Each step is binary-judged for correctness based on alignment with references or logical/visual validity. SAS is computed as the proportion of correct steps.

\paragraph{CoT Reasoning Score (CRS).}
To combine step-wise correctness and final answer validity, we define a weighted reasoning score as \( \text{CRS} = \alpha \cdot \text{SAS} + (1 - \alpha) \cdot \text{Correct}(s_A) \), where \( \text{Correct}(s_A) \in \{0,1\} \) denotes whether the final answer is correct, and \( \alpha \) is set to 0.7 by default.

\paragraph{Key Step Coverage (KSC).}
We also measure the proportion of annotated key reasoning steps that appear in the model output, serving as a recall-style indicator of reasoning completeness.

\subsection{CoT Diversity Evaluation}
\label{subsec:multisolu}

While some recent studies have acknowledged the need for multi-path reasoning evaluation, significant gaps remain.
\cite{zhang2024mathverse} emphasizes that rigid ground-truth templates fail to capture the diversity of reasoning styles, calling for adaptive key-step extraction.
Similarly, \cite{jiang2025mme} and \cite{PhysBench} annotate multiple reasoning paths but lack systematic metrics to measure the ability of models to generate and validate diverse CoT trajectories.

To fill this gap, we introduce \textbf{CoT Diversity Evaluation (CDE)}, a graph-based framework for assessing the ability of models to generate multiple logically valid and distinct reasoning chains.

Specifically, CDE consists of three key stages:

\begin{itemize} [leftmargin=4mm,itemsep=0.03pt,topsep=0.1pt]

    \item \textbf{Reference Graph Construction.} 
    Each annotated instance is converted into directed graphs, with key steps as nodes and logical flows as edges.

    \item \textbf{Model Path Embedding.}  
    We map the model-generated reasoning steps into the reference graph by parsing them into directed edge sequences.

    \item \textbf{Path Matching and Metric Computation.} 
    We define three core metrics for multi-path evaluation:
    \begin{itemize} [leftmargin=4mm,itemsep=0.03pt,topsep=0.1pt]
    
        \item \textbf{Path Validity Rate (PVR)}: Proportion of model edges matching the reference graph.
        \item \textbf{Path Coverage Score (PCS)}: Normalized length of the longest matched sub-path.
        \item \textbf{CoT Match Score (CMS)}: Harmonic mean of PVR and PCS, balancing validity and coverage.
        
    \end{itemize}
\end{itemize}

\paragraph{Path Count Adjustment.}
To fairly compare models with differing numbers of generated and reference paths, we define adjusted versions of the above metrics.

Let \( N_p \) and \( N_{gt} \) denote the numbers of predicted and reference paths, respectively. The adjusted path validity rate is defined as \( \text{Path Validity Rate}_{\text{adj}} = \text{PVR} \times \frac{\min(N_p, N_{gt})}{N_{gt}} \), and the adjusted path coverage score is given by \( \text{Path Coverage Score}_{\text{adj}} = \text{PCS} \times \exp\left(-\alpha \cdot \left(\frac{N_p}{N_{gt}} - 1\right)\right) \), where \( \alpha \) controls the penalty for over-generation: higher values enforce stricter adherence to the reference count, while lower values allow more flexibility.

\begin{table}[t]
\caption{
\textbf{CoT reasoning performance on MVPBench across three dimensions.}
We assess open- and closed-source MLLMs on \textit{CoT Quality} (SAS, KSC, CRS), \textit{CoT Diversity} (PVR, PCS, CMS), and \textit{CoT Efficiency} (SRS, RVR, Avg), under \textit{Single} and \textit{Multi} image settings.
Best single-image results and largest multi-image gains are highlighted for \colorbox{red!10}{closed-source} and \colorbox{blue!10}{open-source} models.
\textcolor{blue}{↑} indicates performance improvement with multi-image input, \textcolor{red}{↓} indicates a drop, and \textcolor{black}{\textasteriskcentered} denotes invalid outputs.
Additional evaluation results for closed-source models and human performance benchmarks are presented in Appendices A.1 and B.1, respectively.
}
\label{tab:cot_evaluation}
\vspace{\baselineskip}
\centering
\tiny
\setlength{\tabcolsep}{1.5pt}
\renewcommand{\arraystretch}{1.15}
\resizebox{\textwidth}{!}{

\begin{tabular}{l|cc cc cc|cc cc cc|cc cc cc}
\toprule
\textbf{Model} & \multicolumn{6}{c|}{\textbf{CoT Quality}} & \multicolumn{6}{c|}{\textbf{CoT Diversity}} & \multicolumn{6}{c}{\textbf{CoT Efficiency}} \\
\cmidrule(lr){2-7} \cmidrule(lr){8-13} \cmidrule(lr){14-19}
& \multicolumn{2}{c}{SAS} & \multicolumn{2}{c}{KSC} & \multicolumn{2}{c|}{CRS} & \multicolumn{2}{c}{PVR} & \multicolumn{2}{c}{PCS} & \multicolumn{2}{c|}{CMS} & \multicolumn{2}{c}{SRS} & \multicolumn{2}{c}{RVR} & \multicolumn{2}{c}{Avg} \\
& Single & Multi & Single & Multi & Single & Multi & Single & Multi & Single & Multi & Single & Multi & Single & Multi & Single & Multi & Single & Multi \\
\midrule
\multicolumn{19}{c}{\textit{\textbf{Open-source MLLMs}}} \\
LLaVA-OV-72B \cite{li2025llavaonevision} & 53.09 & {*} & 29.47 & {*} & 36.49 & {*} & 63.44 & {*} & 70.00 & {*} & 66.98 & {*} & 96.91 & {*} & 99.55 & {*} & 98.23 & {*} \\
LLaVA-CoT \cite{xu2024llava} & 48.47 & \textcolor{blue}{\tiny 8.58↑} & 30.21 & \textcolor{blue}{\tiny 2.23↑} & 32.58 & \textcolor{blue}{\tiny 9.01↑} & 28.87 & \cellcolor{blue!10}\textcolor{blue}{\tiny 10.32↑} & 51.89 & \textcolor{blue}{\tiny 3.75↑} & 34.73 & \cellcolor{blue!10}\textcolor{blue}{\tiny 9.02↑} & 97.63 & \textcolor{red}{\tiny 0.49↓} & 99.64 & \textcolor{blue}{\tiny 0.12↑} & \colorbox{blue!10}{98.64} & \textcolor{red}{\tiny 0.49↓} \\
InternVL2.5-78B \cite{chen2024expanding} & 56.35 & \cellcolor{blue!10}\textcolor{blue}{\tiny 10.12↑} & 42.42 & \textcolor{blue}{\tiny 5.45↑} & 43.98 & \textcolor{blue}{\tiny 4.45↑} & 67.28 & \textcolor{blue}{\tiny 8.43↑} & 72.09 & \textcolor{blue}{\tiny 4.79↑} & 68.72 & \textcolor{blue}{\tiny 5.12↑} & 96.89 & \textcolor{red}{\tiny 0.83↓} & 99.45 & \textcolor{blue}{\tiny 0.50↑} & 98.17 & \textcolor{red}{\tiny 0.16↓} \\
InternVL2.5-78B-MPO \cite{wang2024enhancing} & 55.77 & \textcolor{blue}{\tiny 7.80↑} & 41.87 & \cellcolor{blue!10}\textcolor{blue}{\tiny 5.63↑} & 43.76 & \textcolor{blue}{\tiny 8.51↑} & 72.80 & \textcolor{blue}{\tiny 9.34↑} & 76.08 & \textcolor{blue}{\tiny 5.61↑} & 73.95 & \textcolor{blue}{\tiny 8.11↑} & \colorbox{blue!10}{97.88} & \textcolor{red}{\tiny 1.67↓} & 99.32 & \textcolor{red}{\tiny 0.28↓} & 98.60 & \textcolor{red}{\tiny 0.98↓} \\
InternVL3-78B \cite{InternVL3} & 57.80 & \textcolor{blue}{\tiny 9.26↑} & \colorbox{blue!10}{46.20} & \textcolor{blue}{\tiny 5.49↑} & 47.48 & \cellcolor{blue!10}\textcolor{blue}{\tiny 9.25↑} & 66.06 & \textcolor{blue}{\tiny 7.02↑} & 70.61 & \cellcolor{blue!10}\textcolor{blue}{\tiny 7.53↑} & 67.66 & \textcolor{blue}{\tiny 8.65↑} & 97.54 & \textcolor{blue}{\tiny 0.35↑} & 99.52 & \textcolor{red}{\tiny 0.11↓} & 98.53 & \textcolor{blue}{\tiny 0.13↑} \\
InternVL3-78B-Instruct \cite{InternVL3} & 55.86 & \textcolor{blue}{\tiny 9.53↑} & 42.15 & \textcolor{blue}{\tiny 3.51↑} & 44.24 & \textcolor{blue}{\tiny 8.63↑} & 68.41 & \textcolor{blue}{\tiny 9.78↑} & 72.41 & \textcolor{blue}{\tiny 3.41↑} & 69.81 & \textcolor{blue}{\tiny 8.38↑} & 96.88 & \textcolor{blue}{\tiny 0.29↑} & \colorbox{blue!10}{99.92} & \textcolor{red}{\tiny 0.50↓} & 98.40 & \textcolor{red}{\tiny 0.10↓} \\
Qwen2.5-VL-7B \cite{Qwen2.5-VL} & 52.40 & \textcolor{blue}{\tiny 3.11↑} & 36.54 & \textcolor{blue}{\tiny 1.73↑} & 39.24 & \textcolor{blue}{\tiny 4.32↑} & 64.43 & \textcolor{blue}{\tiny 5.83↑} & 73.70 & \textcolor{blue}{\tiny 2.12↑} & 67.87 & \textcolor{blue}{\tiny 3.86↑} & 93.59 & \textcolor{blue}{\tiny 0.30↑} & 99.26 & \textcolor{red}{\tiny 0.02↓} & 96.43 & \textcolor{blue}{\tiny 0.14↑} \\
Qwen2.5-VL-72B \cite{Qwen2.5-VL} & 57.15 & \textcolor{blue}{\tiny 5.55↑} & 43.29 & \textcolor{blue}{\tiny 5.33↑} & 46.08 & \textcolor{blue}{\tiny 7.24↑} & \colorbox{blue!10}{74.73} & \textcolor{blue}{\tiny 6.76↑} & \colorbox{blue!10}{78.97} & \textcolor{blue}{\tiny 6.12↑} & \colorbox{blue!10}{74.25} & \textcolor{blue}{\tiny 7.34↑} & 97.46 & \textcolor{red}{\tiny 1.50↓} & 99.43 & \textcolor{blue}{\tiny 0.24↑} & 98.45 & \textcolor{red}{\tiny 0.63↓} \\
QVQ-72B \cite{qwen2024qvq} & \cellcolor{blue!10}{68.28} & \textcolor{blue}{\tiny 2.49↑} & 44.63 & \textcolor{red}{\tiny 0.76↓} & \colorbox{blue!10}{53.83} & \textcolor{red}{\tiny 0.88↓} & {*} & {*} & {*} & {*} & {*} & {*} & 85.29 & \cellcolor{blue!10}\textcolor{blue}{\tiny 3.82↑} & 56.27 & \cellcolor{blue!10}\textcolor{blue}{\tiny 3.04↑} & 70.93 & \cellcolor{blue!10}\textcolor{blue}{\tiny 3.28↑} \\
\midrule
\multicolumn{19}{c}{\textit{\textbf{Closed-source MLLMs}}} \\
GPT-4o \cite{openai2024gpt4o} & 63.26 & \cellcolor{red!10}\textcolor{blue}{\tiny 20.30↑} & 46.39 & \cellcolor{red!10}\textcolor{blue}{\tiny 14.75↑} & 50.45 & \cellcolor{red!10}\textcolor{blue}{\tiny 21.41↑} & 68.04 & \cellcolor{red!10}\textcolor{blue}{\tiny 13.22↑} & 72.38 & \textcolor{blue}{\tiny 10.01↑} & 69.50 & \textcolor{blue}{\tiny 13.04↑} & 98.42 & \textcolor{red}{\tiny 1.26↓} & 99.39 & \textcolor{blue}{\tiny 0.28↑} & 98.90 & \textcolor{red}{\tiny 0.49↓} \\
OpenAI o3 \cite{openai2025gpto3} & \colorbox{red!10}{75.29} & \textcolor{blue}{\tiny 15.87↑} & \colorbox{red!10}{50.64} & \textcolor{blue}{\tiny 11.52↑} & \colorbox{red!10}{59.11} & \textcolor{blue}{\tiny 15.83↑} & 68.85 & \textcolor{blue}{\tiny 9.81↑} & 74.91 & \textcolor{blue}{\tiny 10.24↑} & 71.63 & \textcolor{blue}{\tiny 9.97↑} & \colorbox{red!10}{99.43} & \textcolor{red}{\tiny 2.31↓} & \colorbox{red!10}{99.52} & \textcolor{blue}{\tiny 0.13↑} & \colorbox{red!10}{99.48} & \textcolor{red}{\tiny 1.09↓} \\
Claude 3.7 Sonnet \cite{claude3.7} & 64.41 & \textcolor{blue}{\tiny 16.12↑} & 45.66 & \textcolor{blue}{\tiny 11.95↑} & 50.87 & \textcolor{blue}{\tiny 15.22↑} & \colorbox{red!10}{73.70} & \textcolor{blue}{\tiny 12.81↑} & \colorbox{red!10}{75.79} & \cellcolor{red!10}\textcolor{blue}{\tiny 12.04↑} & \colorbox{red!10}{74.25} & \cellcolor{red!10}\textcolor{blue}{\tiny 13.38↑} & 97.76 & \cellcolor{red!10}\textcolor{blue}{\tiny 0.13↑} & 97.34 & \cellcolor{red!10}\textcolor{blue}{\tiny 2.23↑} & 97.55 & \cellcolor{red!10}\textcolor{blue}{\tiny 1.18↑} \\
\bottomrule
\end{tabular}
}
\end{table}

\subsection{CoT Efficiency Evaluation}
\label{subsec:reflection}

The efficiency of reasoning is also crucial for evaluating CoT quality.
Models like o1 generate excessively long reasoning chains with extensive reflection and verification steps.
To capture this aspect, we evaluate the relevance of reasoning steps and the validity of reflective ones.

\paragraph{Step Relevance Score (SRS).}
While long reasoning sequences enable deeper analysis, they often include irrelevant descriptions unrelated to solving the task.  
We partition the model's reasoning into steps and instruct GPT-4o to identify all relevant steps \( P_{\text{relevant}} \).  
A step is considered relevant if its major content directly contributes to problem-solving.  
SRS, similar to SCS, is defined as the proportion of relevant steps among all generated steps.

\paragraph{Reflection Validity Rate (RVR).}
Reflective reasoning can strengthen CoT performance by identifying errors or providing additional justification, but not all reflections are helpful—some may be redundant or incorrect.
We define a reflection step as valid if it (i) identifies a previous error or (ii) offers new supporting reasoning.  
Reflection quality is then measured as the proportion of valid reflections \( R_{\text{valid}} \), detected through linguistic cues such as ``Wait'' or ``Alternatively''.


\section{Comprehensive Evaluation of Multimodal Reasoning via CoT Metrics}

We evaluate various MLLMs using our proposed CoT evaluation suite, with results summarized in Table~\ref{tab:cot_evaluation} and Table~\ref{tab:cot_evaluation_mvpbench}. 
We begin by analyzing the overall performance and then highlight key findings.

\paragraph{Overall Results.}
Table~\ref{tab:cot_evaluation} reports model performance across three CoT evaluation dimensions using SAS, KSC, and SRS for both logical inference and image captioning. Diversity is assessed via PVR and RCS, and robustness is measured by averaging SRS and RVR, with RVR set to 100 for models lacking reflection ability. Table~\ref{tab:cot_evaluation_mvpbench} complements this by presenting subcategory-level evaluation across all CoT metrics on MVPBench.
Model and setup details are in Appendix H.

GPT-4o demonstrates strong overall performance, while OpenAI o3 surpasses it in quality and efficiency, achieving the highest scores. Among open-source models, the InternVL series is most competitive, with InternVL3-vl-78B and MPO-tuned InternVL2.5 showing strong performance across all dimensions. QVQ performs well in CoT quality but lacks robustness, often producing verbose and loosely related content, from which we derive the following key observations.

\paragraph{\textit{CoT Diversity Does Not Guarantee High Reasoning Accuracy.}}
While diversity helps explore multiple reasoning paths, our results show it does not inherently improve reasoning quality. For example, Qwen2.5-VL-72B achieves the highest diversity but underperforms QVQ-72B in quality, despite the latter lacking diversity evaluation. This suggests a trade-off: greater diversity may lead to less focused or accurate reasoning if not properly guided. In contrast, OpenAI o3 attains top quality with moderate diversity, highlighting the importance of goal-directed reasoning.

\paragraph{\textit{Models with reflection largely benefit CoT quality.}}
As shown in Table~\ref{tab:cot_evaluation}, the CRS of QVQ with reflection capability most closely approach GPT-4o.
After specifically fine-tuning for the reasoning capabilities from Qwen2.5-VL-72B, QVQ surpasses its base model by 7.75\%.
Notably, although QVQ generates longer CoT sequences than Qwen2.5-VL-72B, SAS of QVQ still exceeds Qwen2.5-VL-72B by 11.13\%, indicating superior accuracy in each reasoning step.

\paragraph{\textit{Long CoT Models May Be More Prone to Distraction.}}
Models generating longer CoT tend to exhibit lower relevance, often producing content unrelated to the question, reflected by lower KSC scores (compared to QVQ). Some short-CoT models like LLaVA-OV-72B also show low relevance, usually due to repetitive outputs on specific question types. Fine-grained analysis shows models often lose focus when describing images, generating exhaustive but irrelevant captions.

\paragraph{\textit{Reflection Validity Rate often fails to help.}}
While reflection is a key feature of long CoT models for answer verification, QVQ achieve reflection quality scores of only about 56\%, indicating that approximately 44\% of reflection attempts fail to contribute meaningfully to answer accuracy.
This substantial failure rate compromises efficiency by potentially introducing unnecessary or distracting steps before reaching correct solutions.
Future research should explore methods to reduce these ineffective reflections to improve both efficiency and quality.

\paragraph{\textit{Post-training may harm generalization.}}
While post-training—particularly mixed preference optimization (MPO)—is frequently employed to align models more closely with specific downstream tasks, it does not universally enhance CoT reasoning quality. 
As in Figure~\ref{fig:post-training_comparison}, InternVL2.5-78B-MPO underperforms its base counterpart InternVL2.5-78B,  and InternVL3-78B similarly trails  InternVL3-78B-Instruct in Physics Experiments subset.
Although MPO can effectively boost performance on human-preference-aligned subsets such as physics questions, it tends to negatively impact subsets requiring stronger visual perception or temporal prediction capabilities. 
This phenomenon suggests that MPO may introduce distributional biases or lead to overfitting to specific tasks, thereby compromising generalization, visual grounding, and multimodal coherence—particularly evident in visual-centric reasoning tasks. 
MVPBench, with its comprehensive and balanced design across multiple reasoning categories, effectively highlights these limitations.

\section{Understanding the Evaluative Power of \benchmark}
\label{exp}

\begin{table}[t]
\caption{
\textbf{Subcategory-level evaluation of CoT reasoning in MVPBench.}
We present subcategory-level scores for three core reasoning dimensions—\textit{Quality}, \textit{Diversity}, and \textit{Efficiency}—evaluated across both open- and closed-source MLLMs.
Top-performing models within each category are highlighted in \colorbox{blue!10}{blue} (open-source) and \colorbox{red!10}{red} (closed-source).
}
\label{tab:cot_evaluation_mvpbench}
\vspace{\baselineskip}
\centering
\scriptsize
\setlength{\tabcolsep}{1.7pt}
\renewcommand{\arraystretch}{1.05}
\resizebox{\textwidth}{!}{
\begin{tabular}{l|ccc|ccc|ccc|ccc}
\toprule
\textbf{Model} & \multicolumn{3}{c|}{\textbf{Phys-Experiment}} & \multicolumn{3}{c|}{\textbf{Phys-Problems}} & \multicolumn{3}{c|}{\textbf{Spatial-Relation}} & \multicolumn{3}{c}{\textbf{Dyn-Prediction}} \\
& Quality & Diversity & Efficiency & Quality & Diversity & Efficiency & Quality & Diversity & Efficiency & Quality & Diversity & Efficiency \\
\midrule
\multicolumn{13}{c}{\textit{\textbf{Open-source MLLMs}}} \\
LLaVA-OV-72B \cite{li2025llavaonevision} & 37.21 & 56.34 & 94.77 & 32.94 & \cellcolor{blue!10}75.54 & 99.05 & 34.16 & 49.32 & 99.36 & 41.66 & 86.72 & 99.72 \\
LLaVA-CoT \cite{xu2024llava}& 33.79 & 39.41 & \cellcolor{blue!10}97.35 & 20.86 & 34.72 & 98.97 & 31.89 & 39.67 & 98.45 & 43.77 & 25.13 & 99.78 \\
InternVL2.5-78B \cite{chen2024expanding}& 43.95 & 65.06 & 94.25 & 47.44 & 64.27 & 98.83 & \cellcolor{blue!10}39.75 & 62.29 & \cellcolor{blue!10}99.59 & 44.78 & 83.26 &  \cellcolor{blue!10}100 \\
InternVL2.5-78B-MPO \cite{wang2024enhancing}& 41.60 & 73.17 & 97.19 & 51.54 & 72.69 & 98.97 & 37.83 & \cellcolor{blue!10}62.61 & 98.48 & 44.06 & 87.32 & 99.76 \\
InternVL3-78B \cite{InternVL3}& 37.00 & 74.66 & 91.49 & 58.26 & 62.32 &98.92 & 39.31 & 61.29 & 99.14 & 46.68 & 83.88 & 99.95 \\
InternVL3-78B-Instruct \cite{InternVL3}& 42.01 & 66.91 & 94.87 & 52.64 & 64.50 & \cellcolor{blue!10} 99.81 & 38.10 & 61.89 & 98.96 & 44.20 & 85.94 & 99.96 \\
Qwen2.5-VL-7B \cite{Qwen2.5-VL}& 37.00 & 74.66 & 91.49 & 42.34 & 63.60 & 98.55 & 35.20 & 54.70 & 95.82 & 40.30 & 78.50 & 99.85 \\
Qwen2.5-VL-72B \cite{Qwen2.5-VL}& 41.19 & \cellcolor{blue!10}77.15 & 96.72 & 57.01 & 73.82 & 99.36 & 39.18 & 47.70 & 98.06 & 46.94 & \cellcolor{blue!10}98.32 & 99.65 \\
QVQ-72B \cite{qwen2024qvq} & \cellcolor{blue!10}49.63 & 0.00 & 71.65 & \cellcolor{blue!10}60.97 & 0.00 & 63.71 & 38.50 & 0.00 & 69.24 & \cellcolor{blue!10}66.20 & 0.00 & 79.13 \\
\midrule
\multicolumn{13}{c}{\textit{\textbf{Closed-source MLLMs}}} \\
GPT-4o \cite{openai2024gpt4o}& 50.21 & 65.73 & \cellcolor{red!10}97.53 & 52.29 & 63.55 & 98.77 & 43.64 & 60.17 & 99.72 & 52.35 & 88.76 & 99.59 \\
OpenAI o3 \cite{openai2025gpto3}& \cellcolor{red!10}57.73 & 68.36 & 97.44 &\cellcolor{red!10}65.36 & 64.30 & \cellcolor{red!10}99.06 & \cellcolor{red!10}43.92 & \cellcolor{red!10}66.11 & \cellcolor{red!10}99.83 & \cellcolor{red!10}69.44 & 87.75 & \cellcolor{red!10}99.71 \\
Claude 3.7 Sonnet \cite{claude3.7}& 49.13 & \cellcolor{red!10}74.20 & 97.38 & 57.02 & \cellcolor{red!10}68.53 & 94.71 & 42.41 & 61.96 & 99.67 & 54.92 & \cellcolor{red!10}92.29 & 98.45 \\
\bottomrule
\end{tabular}
}
\end{table}

Our dataset, \benchmark, is specifically constructed to test multimodal reasoning under diverse and fine-grained physical scenarios.
We explore its impact on evaluation outcomes from two perspectives: category diversity and input modality. 

\paragraph{\textit{Category diversity influences evaluation difficulty.}}
\benchmark\ spans a variety of physical reasoning subcategories, each posing distinct challenges. We observe that model performance varies substantially across these categories, underscoring the impact of task type on evaluation difficulty. For example, InternVL3-78B achieves a Quality score of 58.26 on the more abstract \textit{Phys-Problems} category, but performs better with a score of 66.68 on the more concrete \textit{Dyn-Prediction} tasks (see Table~\ref{tab:cot_evaluation_mvpbench}). Notably, across all open-source models, the \textit{Spatial-Relation} subset yields the lowest average Quality score (37.10), suggesting it poses the greatest challenge. This indicates that current MLLMs still struggle with fine-grained spatial reasoning, revealing a critical gap in their perceptual and relational understanding of physical scenes. This performance gap illustrates how reasoning complexity varies by category and highlights the importance of category-aware evaluation for robust and meaningful model comparisons.

\begin{figure*}[t]
  \centering
  \begin{minipage}[c]{0.42\textwidth}
    \centering
    \includegraphics[width=\linewidth]{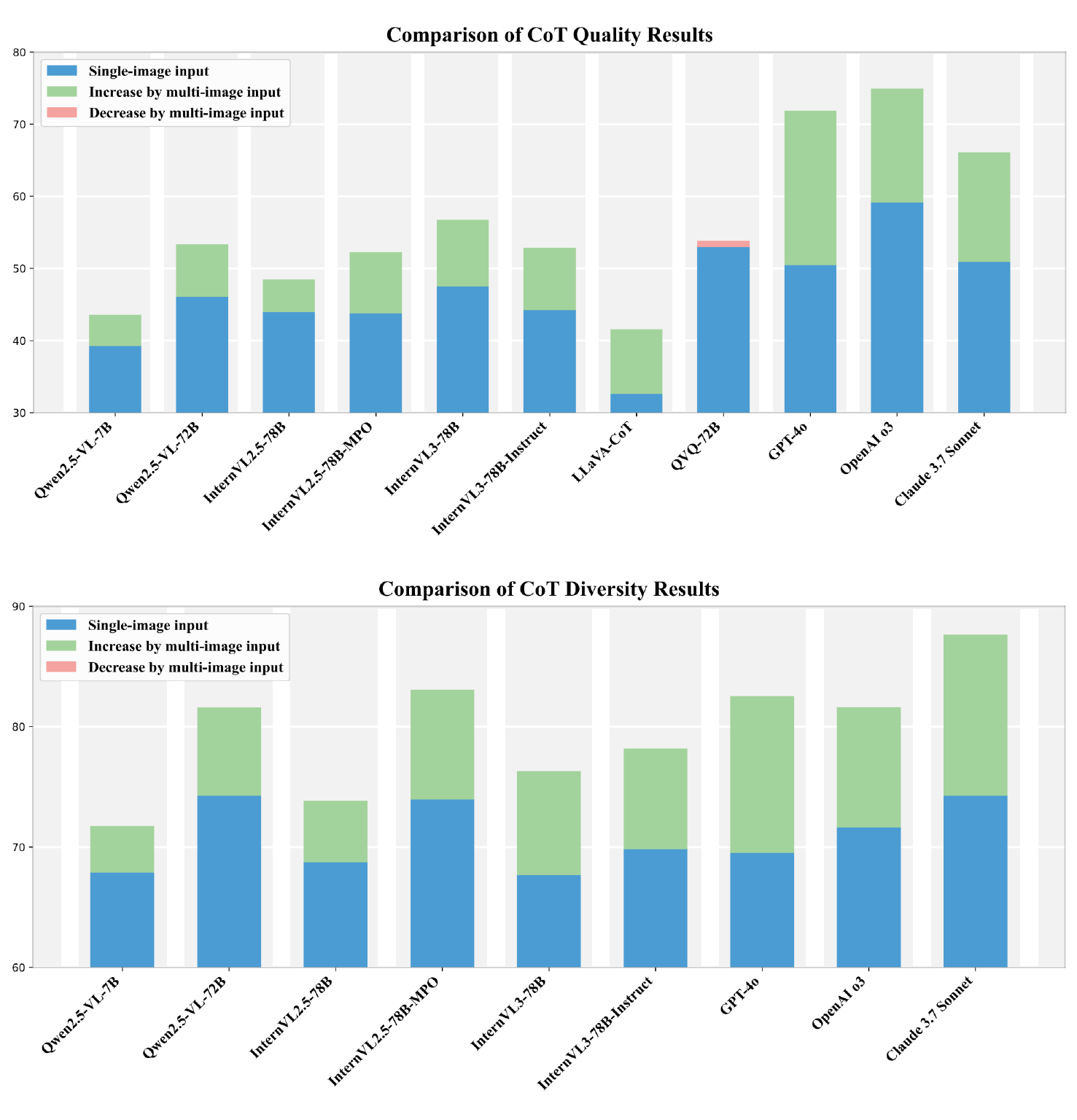}
    \vspace{-15pt} 
    \captionof{figure}{%
      \textbf{Performance comparison between single-image and multi-image inputs on CoT evaluation.}
      This figure highlights the performance difference when reasoning with multiple images versus a single image across various MLLMs. 
      Multi-image inputs generally enhance performance, while QVQ shows a drop—indicating potential challenges in multi-image integration.
    }
    \label{fig:inputs_comparision}
  \end{minipage}\hfill
  \begin{minipage}[c]{0.53\textwidth}
    \centering
    \includegraphics[width=\linewidth]{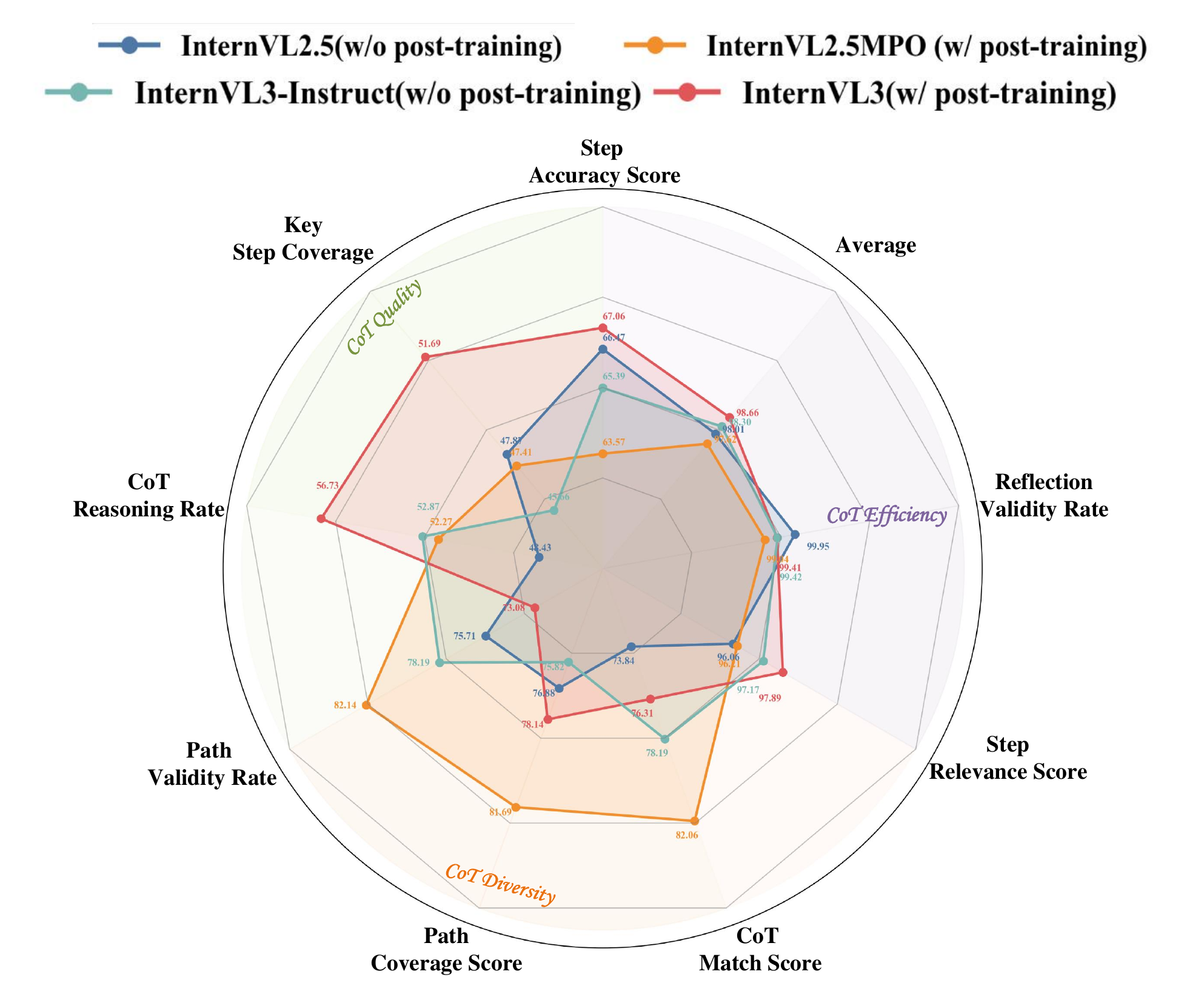}
    \vspace{-15pt} 
    \captionof{figure}{%
      \textbf{CoT Performance of MLLMs with post-training versus without post-training.}  
      InternVL2.5 and InternVL3-instruct represent models without post-training, whereas InternVL2.5-MPO and InternVL3 denote their post-trained counterparts.
      Please note that each metric axis has its own independent scale.
      The results clearly indicate that post-training often fails to enhance the reasoning performance of models and degrades it.
    }
    \label{fig:post-training_comparison}
  \end{minipage}
\end{figure*}

\paragraph{\textit{Multi-image input significantly boosts model performance.}}
To evaluate the impact of input modality, we conducted comparative experiments using both single-image and multi-image inputs under identical prompts and evaluation metrics. This design isolates the effect of visual input quantity, allowing for a controlled analysis of performance variance. As illustrated in Fig~\ref{fig:inputs_comparision}, nearly all models benefit from multi-image inputs, achieving notable gains in both CoT Quality and Diversity scores. Closed-source models show particularly striking improvements, with GPT-4o leading the trend—its CoT Quality score rises from 50 to 72, a relative increase of 44\%, and its Diversity score jumps from 70 to 85, a 21\% improvement. Other closed-source models like Claude 3.7 Sonnet and OpenAI o3 also exhibit significant gains, with Quality scores increasing by 15\% and Diversity by 13\%. Open-source models, such as InternVL3-78B, show more modest improvements, rising from a Quality score of 47.5 to 56.7 (a 19\% increase) and a Diversity score improvement of around 10\%. However, QVQ-72B is an outlier, showing a performance drop of roughly 1-2 points in quality, indicating potential challenges in multi-image integration. Overall, these results highlight the superior adaptability of closed-source models, particularly GPT-4o, in leveraging multi-image inputs to enhance fine-grained physical reasoning and diversity in responses.

\section{Conclusion}
We introduce \benchmark, a benchmark designed to rigorously evaluate visual chain-of-thought reasoning in multimodal large language models (MLLMs). 
It target tasks that require grounded, multi-step inference over visual evidence and goes beyond surface-level image description.
Our evaluation reveals than even state-of-the-art models like GPT-4o and OpenAI o3  often struggle with physical reasoning.
To diagnose these failures, we introduce a graph-based CoT consistency metric to assess reasoning validity, uncovering frequent violations of basic physical principles. 
Notably, we find that reinforcement learning-based alignment can impair physical reasoning, highlighting a misalignment between current fine-tuning strategies and the demands of physical perceptual reasoning. 
These findings  call for post-training strategies that better integrate  visual grounding, causal structure, and structured explananation in MLLMs.


{\small
\bibliographystyle{unsrt}
\bibliography{main}
}





\appendix

\section*{Appendix Overview}

Our supplementary includes the following sections:

\begin{itemize}
  \item \textbf{Section \ref{AppendA}: More experiment results.} Extended Empirical Analysis on Closed-source and Post-trained Models.
  \item \textbf{Section \ref{AppendB}: More Exploration.} Analysis of human performance and error analysis.
  \item \textbf{Section \ref{AppendC}: More Dataset Details.}
  \item \textbf{Section \ref{AppendD}: More Qualitative Examples.} More visualization of our evaluation demos.
  \item \textbf{Section \ref{AppendE}: Limitations.} Discussion of limitations of our work.
  \item \textbf{Section \ref{AppendF}: Broader impacts.} Discussion of societal impacts of our work.
  \item \textbf{Section \ref{AppendG}: Detailed Evaluation prompts.} 
  \item \textbf{Section \ref{AppendH}: Setup.} Details for model design, implementation.
\end{itemize}

We have shared the following artifacts:

\begin{table}[h!]
\centering
\begin{tabular}{@{}ll@{}}
\toprule
\textbf{Artifact} & \textbf{Link} \\
\midrule
Code Repository & \href{https://github.com/CSU-JPG/MVPBench}{https://github.com/CSU-JPG/MVPBench} \\
Data & \href{=https://huggingface.co/datasets/CSU-JPG/MVPBench}{https://huggingface.co/datasets/CSU-JPG/MVPBench} \\
\bottomrule
\end{tabular}
\end{table}

The authors are committed to ensuring its regular upkeep and updates.

\newpage

\section{More experiment results}
\label{AppendA}

\subsection{More Closed-source Model Experiments}

\begin{table}[H]
\caption{
\textbf{Additional Evaluation Results for Closed-Source Models on CoT Reasoning Performance across Three Dimensions in MVPBench.}
\textcolor{blue}{↑} indicates performance improvement with multi-image input, \textcolor{red}{↓} indicates a drop.
}
\label{tab:closed}
\vspace{\baselineskip}
\centering
\tiny
\setlength{\tabcolsep}{1.5pt}
\renewcommand{\arraystretch}{1.15}
\resizebox{\textwidth}{!}{

\begin{tabular}{l|cc cc cc|cc cc cc|cc cc cc}
\toprule
\textbf{Model} & \multicolumn{6}{c|}{\textbf{CoT Quality}} & \multicolumn{6}{c|}{\textbf{CoT Diversity}} & \multicolumn{6}{c}{\textbf{CoT Efficiency}} \\
\cmidrule(lr){2-7} \cmidrule(lr){8-13} \cmidrule(lr){14-19}
& \multicolumn{2}{c}{SAS} & \multicolumn{2}{c}{KSC} & \multicolumn{2}{c|}{CRS} & \multicolumn{2}{c}{PVR} & \multicolumn{2}{c}{PCS} & \multicolumn{2}{c|}{CMS} & \multicolumn{2}{c}{SRS} & \multicolumn{2}{c}{RVR} & \multicolumn{2}{c}{Avg} \\
& Single & Multi & Single & Multi & Single & Multi & Single & Multi & Single & Multi & Single & Multi & Single & Multi & Single & Multi & Single & Multi \\
\midrule
\multicolumn{19}{c}{\textit{\textbf{Closed-source MLLMs}}} \\
Gemini-2.5-flash-preview-04-17 \cite{geminiteam2024geminifamilyhighlycapable} & 60.56 & \textcolor{blue}{\tiny 12.32↑} & 49.29 & \textcolor{blue}{\tiny 8.54↑} & 50.05 & \textcolor{blue}{\tiny 11.63↑} & 56.44 & \textcolor{blue}{\tiny 9.23↑} & 59.35 & \textcolor{blue}{\tiny 7.12↑} & 57.20 & \textcolor{blue}{\tiny 8.45↑} & 97.59. & \textcolor{red}{\tiny 0.37↓} & 92.00 & \textcolor{blue}{\tiny 2.00↑} & 94.71 & \textcolor{blue}{\tiny 0.82↑} \\
Grok3\cite{Grok3}  & 62.48 & \textcolor{blue}{\tiny 3.44↑} & 52.05 & \textcolor{blue}{\tiny 4.13↑} & 52.69 & \textcolor{blue}{\tiny 4.50↑} & 61.57 & \textcolor{blue}{\tiny 10.13↑} & 68.05 & \textcolor{blue}{\tiny 6.89↑} & 63.78 & \textcolor{blue}{\tiny 8.43↑} & 89.55 & \textcolor{red}{\tiny 2.53↓} & 86.00 & \textcolor{blue}{\tiny 6.26↑} & 87.77 & \textcolor{blue}{\tiny 1.87↑} \\
\bottomrule
\end{tabular}
}
\end{table}

To evaluate additional closed-source models, we randomly sampled 25 instances from each sub-dataset of MVPBench, resulting in 100 samples in total. As shown in Table~\ref{tab:closed} and Table~\ref{tab:sub}, the results of these models largely confirm the trends observed with tested models discussed earlier: performance varies notably across different sub-datasets, and multi-image input consistently leads to substantial improvements. Interestingly, Gemini~\cite{geminiteam2024geminifamilyhighlycapable} demonstrates strong quality in the \textit{Physics Experiments} subset, yet performs surprisingly poorly in the \textit{Spatial Relations} task—even falling behind several open-source models.

\subsection{More Post-training Model Experiments}
\begin{figure*}[htbp]
    \centering
    \includegraphics[width=0.8\linewidth]{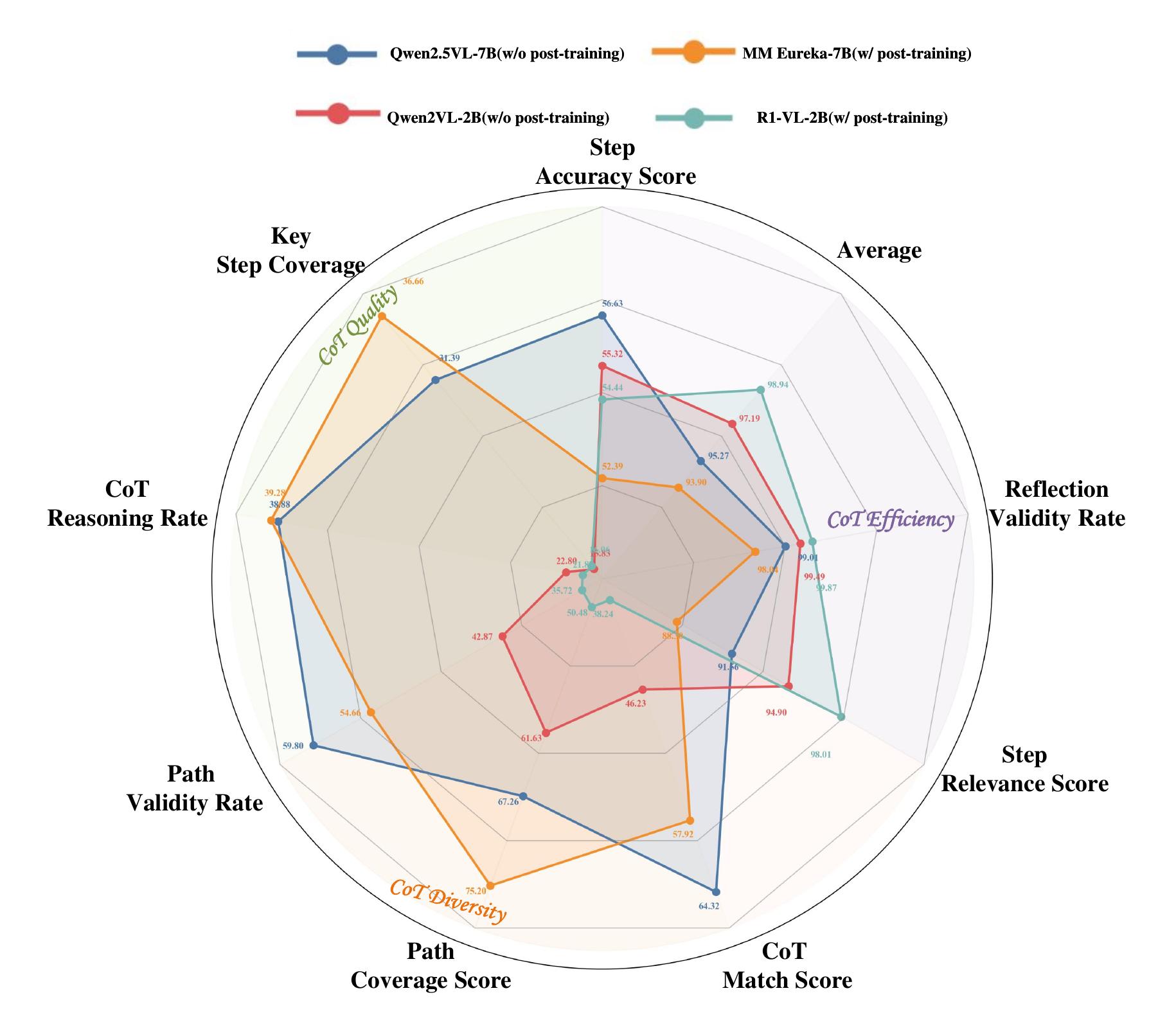}
    \caption{
    \textbf{CoT Performance of MLLMs with post-training versus without post-training.}  
    Qwen2.5VL-7B and Qwen2VL-2B represent models without post-training, whereas MM Eureka-7B and R1-VL-2B denote their post-trained counterparts.Please note that each metric axis has its own independent scale.The results clearly indicate that post-training fails to enhance the reasoning performance of models and degrades it.
    }
    \label{fig:more_post-training_comparison}
\end{figure*}

To further investigate the impact of post-training on model generalization, we conducted additional experiments comparing different base models and distinct post-training methods. 
Specifically, we compared two base models without post-training, Qwen2.5VL-7B and Qwen2VL-2B, against their respective post-trained counterparts: MM Eureka-7B, which employs large-scale rule-based reinforcement learning (RL), and R1-VL-2B, utilizing Step-wise Group Relative Policy Optimization (StepGRPO). 
The comparative analysis indicates clear trends consistent with our earlier findings in the InternVL series. 
As shown in Figure \ref{fig:more_post-training_comparison}, Qwen2.5VL-7B exhibits superior Step Accuracy (56.63\%) compared to MM Eureka-7B (52.39\%). Similarly, Qwen2VL-2B outperforms R1-VL-2B in Path Validity Rate (42.87\% versus 35.72\%) and Path Coverage Score (61.63\% versus 50.48\%), demonstrating significant performance drops associated with post-training methods.
Although certain metrics like Key Step Coverage show modest improvements in post-trained models (MM Eureka-7B: 36.66\% vs. Qwen2.5VL-7B: 31.39\%), the overall pattern emphasizes a general reduction in multimodal coherence and visual-centric reasoning effectiveness post-training.
These findings align with observations from the InternVL models discussed in the main text and reinforce the conclusion that various post-training approaches, despite improving alignment to specific tasks, may impair generalization, particularly in visual-centric and dynamic reasoning tasks.

\section{More Exploration}
\label{AppendB}

\subsection{Human Performance}

To estimate human performance, we recruited four undergraduate students who had received systematic training in physics and were familiar with fundamental physical concepts. Each student was asked to solve the same 100 instances used in our closed-source model evaluation. Unlike other benchmarks, MVPBench is formulated as a visual question answering (VQA) task, and the evaluation of \textit{quality} and \textit{efficiency} relies on the generation of detailed, step-by-step reasoning chains. Therefore, our human performance assessment focuses solely on the \textit{diversity} metric. For each instance, students were provided with the question, answer, image(s), and annotated key reasoning steps. They were instructed to produce as many distinct reasoning chains as possible that could lead to the correct answer by covering all the provided key steps. The resulting outputs were then used to compute the diversity scores.

\begin{table}[H]
\caption{
\textbf{Expanded Subcategory-level Evaluation of CoT Reasoning in MVPBench: Closed-Source Models and Human Baselines.}
We present a detailed subcategory-level evaluation of CoT reasoning along the dimensions of \textit{Quality}, \textit{Diversity}, and \textit{Efficiency}, comparing closed-source MLLMs with human performance on MVPBench.
}
\label{tab:sub}
\vspace{\baselineskip}
\centering
\scriptsize
\setlength{\tabcolsep}{1.7pt}
\renewcommand{\arraystretch}{1.05}
\resizebox{\textwidth}{!}{
\begin{tabular}{l|ccc|ccc|ccc|ccc}
\toprule
\textbf{Model} & \multicolumn{3}{c|}{\textbf{Phys-Experiment}} & \multicolumn{3}{c|}{\textbf{Phys-Problems}} & \multicolumn{3}{c|}{\textbf{Spatial-Relation}} & \multicolumn{3}{c}{\textbf{Dyn-Prediction}} \\
& Quality & Diversity & Efficiency & Quality & Diversity & Efficiency & Quality & Diversity & Efficiency & Quality & Diversity & Efficiency \\
\midrule
\multicolumn{13}{c}{\textit{\textbf{Human Performance}}} \\
& - & 98.72 & - & - & 96.42 & - & - & 99.13 & - & - & 95.76 & - \\
\midrule
\multicolumn{13}{c}{\textit{\textbf{Closed-source MLLMs}}} \\
Gemini-2.5-flash-preview-04-17 \cite{geminiteam2024geminifamilyhighlycapable}& 61.85
 & 68.64 & 100.00 & 63.37 & 39.10 & 85.56 & 28.36 & 73.04 & 93.26 & 46.62 & 48.00 & 100.00 \\
Grok3\cite{Grok3} & 43.85 & 65.54 & 87.60 & 58.16 & 72.26 & 78.50 & 50.60 & 58.72 & 85.66 & 58.16 & 58.59 & 99.33 \\
\bottomrule
\end{tabular}
}
\end{table}

\subsection{Error Analysis}

\begin{figure*}[t]
    \centering
    \includegraphics[width=\linewidth]{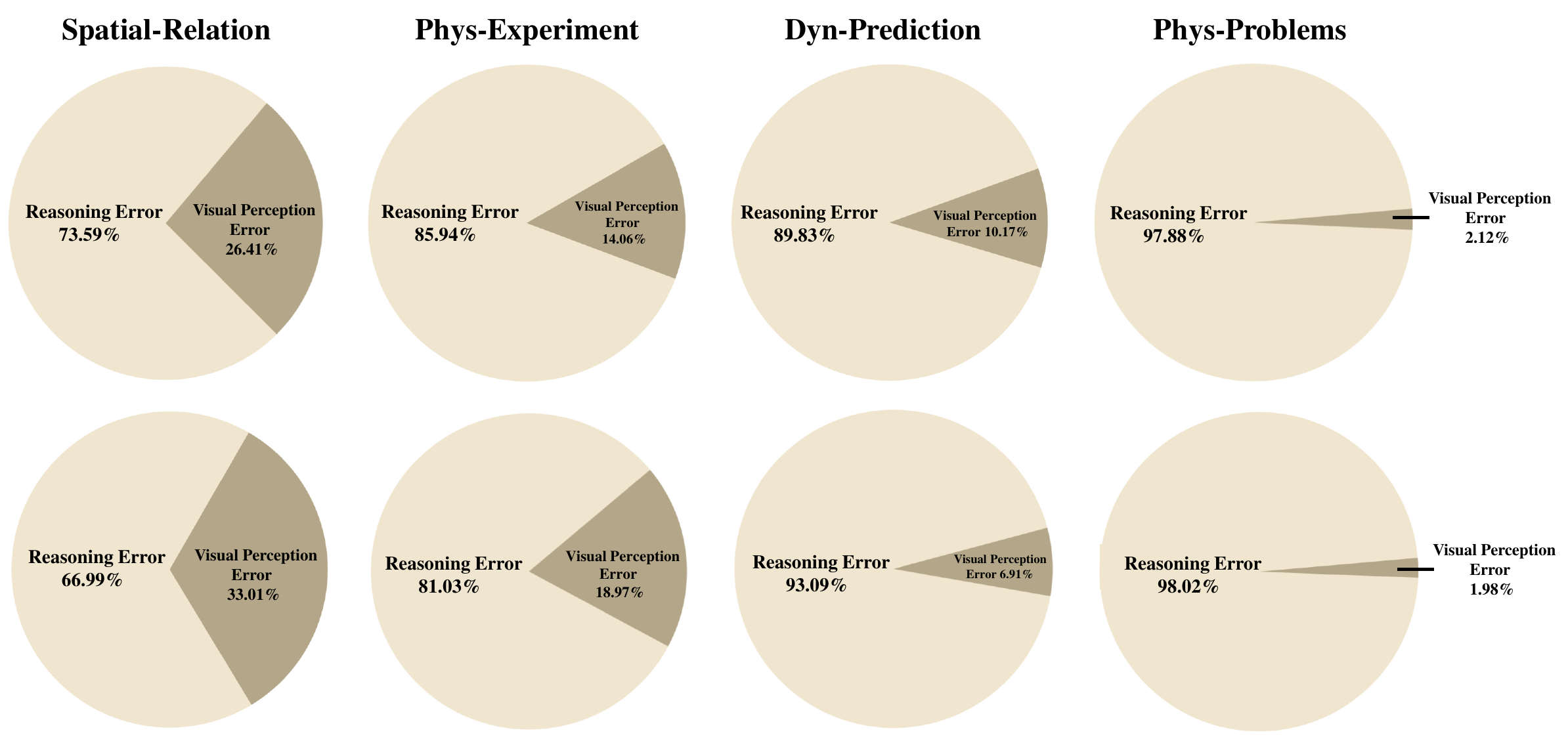}
    \caption{
    \textbf{Distribution of GPT-4o \cite{openai2024gpt4o} Errors across Different Types.} 
    We report the error distribution of GPT-4o on MVPBench, categorized into two types: \textit{Visual Perception Errors} and \textit{Reasoning Errors}, across four representative subcategories. 
    The first row illustrates the error distribution under single-image input settings, while the second row presents results under multi-image inputs.
    }
    \label{fig:error_analysis}
\end{figure*}

To delve into the fine-grained predictions, we select the best-performing MLLM, GPT-4o\cite{openai2024gpt4o}, to understand its modes of success and failure. Our proposed CoT evaluation strategy has produced a detailed assessment of model output, including step-wise scores and explanation, reducing extensive manual effort in identifying and analyzing errors. As shown in Figure~\ref{fig:error_analysis}, we conduct our analysis on the two-step output from the CoT evaluation across the entire dataset, focusing on two key dimensions.

\paragraph{Reasoning Errors Dominate Across Subcategories.}
In particular, the proportion of visual perception errors in the physics-related subset is remarkably low—only 2.12\% and 1.98\% under single- and multi-image inputs, respectively. This finding contrasts with prior observations in MathVerse \cite{zhang2024mathverse}, highlighting the distinct characteristics of our benchmark. We posit that, within our dataset, GPT-4o is generally able to perceive the visual input correctly, but often fails during the reasoning process, leading to incorrect final answers.

\paragraph{Spatial-Relation Emerge as a Major Source of Perception Failures.}
In the spatial-relation subset, visual perception errors account for a striking 33.01\% and 26.41\% under single- and multi-image settings, respectively—substantially higher than in other subsets. This aligns with earlier findings that both closed-source and open-source MLLMs consistently perform worst on spatial relation tasks in terms of the quality metric. These results further support our initial hypothesis: current models struggle significantly with visual grounding when interpreting spatial relationships, underscoring a persistent bottleneck in multimodal understanding.

\label{subsec:ana}

\section{More Dataset Details}
\label{AppendC}

\subsection{Data Collection}
To support the evaluation of multimodal physical reasoning, we constructed a diverse and well-structured dataset spanning four distinct subdomains: (1) physics experiment videos, (2) conceptual physics questions, (3) spatial reasoning images, and (4) dynamic physical scene videos.
\textbf{The annotation process was carried out between March 28 and May 14, 2025, by a team of 31 annotators with backgrounds in physics, science education, and computer vision.}
Each data modality followed a carefully designed protocol to ensure quality, consistency, and relevance to downstream reasoning tasks.

\begin{table}[ht]
\centering
\caption{Annotation summary across the four data modalities.}
\label{tab:annotation_stats}
\begin{tabular}{lccc}
\toprule
\textbf{Data Type} & \textbf{Sample Count} & \textbf{Average Length} & \textbf{Annotators} \\
\midrule
Physics Experiment Videos & 440 & ~60 seconds & 16 \\
Conceptual Physics Problems & 320 & ~200 words & 7 \\
Spatial Reasoning Images & 400 & 1 image & 4 \\
Dynamic Scene Videos & 100 & ~2 seconds & 4 \\
\bottomrule
\end{tabular}
\end{table}

\paragraph{Physics Experiment Videos.}
This subset consists of 440 real-world videos sourced primarily from science education creators on Bilibili, such as "Lighthouse Laboratory" and "Interesting physics in life". 
These videos depict demonstrative physics experiments across domains including mechanics, optics, electromagnetism, and thermodynamics.
Each video was segmented into a sequence of 3 to 5 keyframes capturing critical steps of a physical process.
Annotators provided a natural language description for the initial state, intermediate key steps (each with conclusions), and a final outcome.
Visual markers (e.g., arrows, labeled objects) were optionally added to enhance clarity. 
Multiple plausible reasoning chains were manually curated to reflect different logical paths. 
All samples underwent double annotation with inter-annotator agreement checks and periodic expert reviews. The average duration per video was approximately 60 seconds.

\paragraph{Conceptual Physics Problems.}
This subset includes 320 multiple-choice and short-answer physics questions derived from high school curricula and online education platforms. 
Each item was manually adapted to include visual support (e.g., diagrams or plots), and transformed into a question-answer format with structured reasoning chains.
Annotators selected questions where visual content was essential to reasoning, added visual cues to images (e.g., red dots, arrows), and reformulated options into logical deduction steps.
Stepwise reasoning was expressed using Markdown-compatible mathematical expressions to support neural symbolic processing. 
The annotation reference document for this task was "MCoT-phytest.docx."
All data underwent double annotation and review for logical soundness, visual accuracy, and completeness. On average, each problem included ~200 words of reasoning and annotations.

\paragraph{Spatial Reasoning Images.}
This subset comprises 400 images curated from public domain resources such as Unsplash, Pixabay, and Archive.org. 
It addresses four categories of spatial reasoning: directional relations, distance estimation, perspective transformations, and topological connectivity.
Annotators formulated tasks such as "What direction is object A facing?" or "From the first-view perspective of object A, where is object B?", using generic language to avoid lexical leakage. 
Key steps were illustrated using labeled visual cues and blue/red markings. Logical reasoning was written in natural language chains, each step tied to a specific visual cue or interpretation.
Annotation was guided by the document "MCoT-spatial.docx" and performed by 4 annotators with experience in spatial cognition and vision tasks.

\paragraph{Dynamic Physical Scene Videos.}
The final subset includes 100 short video clips (average duration ~2 seconds) selected from the PhysBench dataset. The tasks focus on predicting physical dynamics, such as object collision trajectories, liquid flow directions, and stability outcomes.
Annotators extracted representative keyframes from each video and documented the physical evolution using a minimal chain of reasoning steps. For instance, a liquid falling through barriers would be annotated by highlighting key deflection events and predicting the final compartment of flow. Problems were written in standardized English using referential expressions (e.g., object A, path B).
All dynamic samples followed the procedure detailed in "dynamic-prediction.docx," and were annotated by 4 individuals with expertise in physics simulation and time-series interpretation.
\begin{figure*}[htbp]
    \centering
    \includegraphics[width=\linewidth]{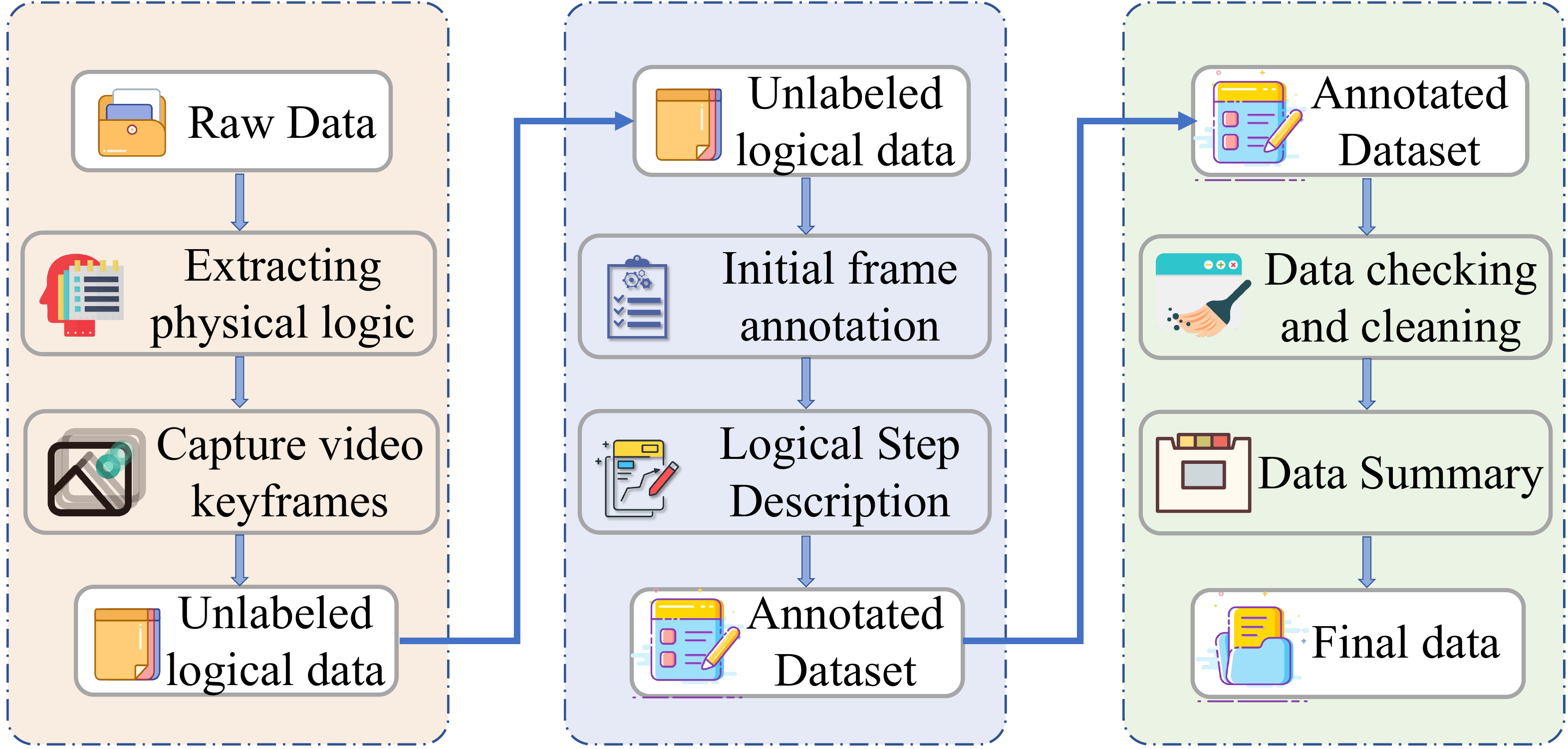}
    \caption{
    \textbf{Data collection process.}
    Initially, all visual and textual data undergo rigorous manual selection to ensure accuracy and relevance. 
    Subsequently, expert annotators manually identify and highlight key objects and events, marking them visually with indicators such as arrows, and provide precise textual annotations for each critical step. 
    Finally, multiple reasoning chains and key step annotations are meticulously constructed and validated manually, ensuring high-quality, reliable data for evaluating multimodal reasoning capabilities.
    }
    \label{fig:motivation}
\end{figure*}

\subsection{Detailed of MVPBench Composition}

\paragraph{Physics Experiments.}
The Physics Experiments subset of MVPBench contains a curated collection of 400 experimental questions, each designed to evaluate a model’s understanding of sequential physical processes through multi-step visual inference. 
These experiments span five fundamental categories: Mechanics (222 questions), Thermodynamics (90 questions), Electromagnetism (42 questions), Optics (33 questions), and Kinematics (13 questions). 
Models must visually interpret the sequence of events and logically deduce the physical processes involved. 
In Mechanics tasks, models must interpret scenarios involving force interactions and motion, whereas Thermodynamics problems require reasoning about heat transfer and energy dynamics. 
Electromagnetism experiments involve interpreting visual representations of electric circuits and magnetic fields.
Optics tasks test understanding of light behavior, reflection, and refraction, while Kinematics scenarios focus on analyzing motion trajectories and velocities. 
These tasks collectively ensure that the evaluated models develop comprehensive visual reasoning abilities similar to how humans mentally simulate physical experiments.

\paragraph{Physics Problems.}
The Physics Problems subset contains a total of 311 challenging, visually grounded physics questions,  primarily sourced from academic examination databases such as Chinses Gaokao physics questions, the International Physics Olympiad (IPhO), and Chinese Mock Examinations at Various Levels, further augmented by additional questions from the PhysReason-mini dataset. 
These problems span five core physics categories: Mechanics (58 questions), Thermodynamics (56 questions), Electromagnetism (90 questions), Optics (53 questions), and Kinematics (54 questions). 
Mechanics questions may involve complex analysis of force interactions or equilibrium scenarios, while Thermodynamics problems often present visual cues related to heat exchange and energy conversion processes. 
Electromagnetism tasks require reasoning about visually depicted electric circuits and magnetic field interactions. 
Optics questions focus on image formation, lens behavior, and optical phenomena, and Kinematics challenges typically demand interpretation of visual trajectories, acceleration, and velocity vectors. 
This detailed structuring and multimodal approach aim to assess models' capabilities in accurately interpreting visual information and applying advanced reasoning to solve intricate physics problems.

\paragraph{Spatial Relations.}
The Spatial Relations subset assesses spatial perception through 400 carefully designed questions, divided into four specific subcategories.
(1) Direction Judgment (100 questions): This subcategory requires models to accurately determine the relative directional positioning of various objects within a scene, emphasizing an understanding of spatial orientation and relational positioning. 
(2) Distance Estimation (100 questions): Tasks here involve estimating the distance and depth relations between objects or between objects and the camera viewpoint, highlighting the importance of accurate depth perception and visual estimation skills. 
(3) First-view Transformation (100 questions): This subcategory challenges models to reason about spatial directions from an egocentric viewpoint, simulating real-world scenarios where orientation judgments are made from a first-person perspective. 
(4) Topological Relation Judgment (100 questions): This category focuses specifically on assessing the reachability and connectivity within directed graphs, using images constructed through graphical editing tools.
Overall, this subset is designed to rigorously evaluate models' capabilities in processing complex spatial scenarios and performing accurate spatial reasoning, reflecting essential cognitive processes used in navigating and interpreting real-world visual environments.

\paragraph{Dynamic Prediction.}
The Dynamic Prediction subset comprises 100 tasks designed to evaluate the predictive capabilities of models regarding dynamically evolving physical interactions, structured into four subcategories: 
(1) Multi-object Collision (25 questions): This category requires models to predict outcomes involving interactions among multiple objects, such as collisions, considering momentum, energy transfer, and motion trajectories.
(2) Liquid Diversion (25 questions): Tasks involve predicting fluid paths through variously configured channels or obstacles, necessitating models to understand fluid dynamics visually.
(3) Physical State Prediction (25 questions): These problems challenge models to anticipate changes in the physical states of objects, such as transitions between solid, liquid, and gas phases, based on visual cues and temporal sequences.
(4) Shadow Transformation Prediction (25 questions): This subcategory assesses the ability of models to predict and interpret the changes in shadows cast by objects due to movements or shifts in light sources, requiring sophisticated temporal and spatial reasoning.
These tasks collectively aim to test models' capacity to interpret and forecast dynamic physical phenomena, thereby closely replicating human cognitive processes involved in visual prediction and temporal reasoning.
\subsection{Data Analysis}
\begin{figure}[htbp]
  \centering
  \begin{minipage}[c]{0.48\textwidth}
    \centering    
    \begin{tabular}{lr}
      \toprule
      \textbf{Statistic}               & \textbf{Value} \\ 
      \midrule
      Total samples                    & \TotalData   \\
      Total images                     & 4,701   \\ 
      Unique images                    & 4,688   \\
      Unique questions                 & 1,211  \\
      Unique answers                   & 1,211   \\
      \midrule
      Max. question length             & 100     \\
      Avg. question length             & 28.01  \\
      Max. answer steps                & 9      \\
      Avg. answer steps                & 2.93   \\
      \midrule
      Max. Image-CoTs per sample             & 5      \\
      Avg. Image-CoTs per sample             & 3.90   \\
      Max. reasoning paths             & 16     \\
      Avg. reasoning paths             & 2.67   \\
      \bottomrule
    \end{tabular}
    \captionof{table}{
    \textbf{Key statistics of MVPBench.}
    Summarizes dataset size, question/answer properties, and multi-path reasoning annotations for evaluating complex reasoning in MLLMs.
    }
    \label{tab:key-stats}
  \end{minipage}%
  \hfill
  \begin{minipage}[c]{0.45\textwidth}
    \centering
    \includegraphics[width=\linewidth]{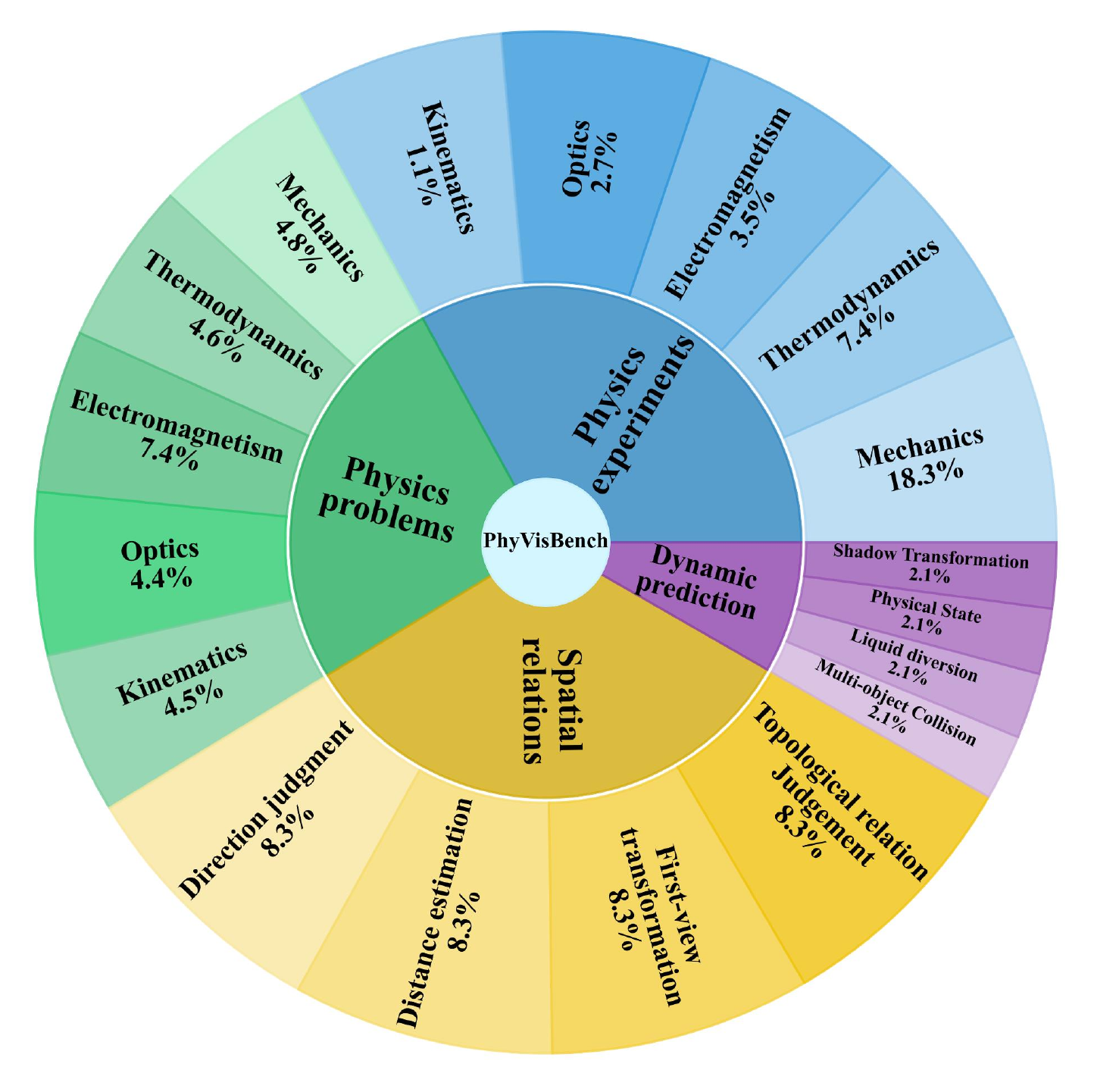}
    \caption{
    \textbf{Category distribution in MVPBench.}
    Covers 4 major reasoning categories and 18 fine-grained subcategories.
    }
    \label{fig:benchmark-distribution}
  \end{minipage}
\end{figure}

Table \ref{tab:key-stats} presents core statistics of the \benchmark\ dataset, which consists of \TotalData samples with a total of 4,701 images, covering both unique and repeated images. 
Each question and corresponding answer is distinct, underscoring the dataset’s broad range and depth across various physical reasoning scenarios.
Furthermore, question lengths display considerable variation, with some reaching up to 100 words, though the majority of questions are moderately sized.
Answers generally involve multiple reasoning steps, reflecting a significant complexity level within the dataset.
Notably, the dataset includes multiple Image-CoTs per sample—visual chains of thought specifically crafted as input to guide and assess model reasoning processes. 
The average number of Image-CoTs per sample is approximately 3.90, with some samples containing up to 5, ensuring rich visual context for enhanced multimodal reasoning. 
Additionally, each sample captures several chains of thought, facilitating the evaluation of multi-path reasoning capabilities. 

The dataset includes multiple subsets(Figure \ref{fig:benchmark-distribution}), with Physics Experiments and Spatial Relations forming the most significant components, emphasizing sequential reasoning through multi-step physical processes and complex spatial perception tasks, respectively.
Additionally, a substantial contribution from the Physics Problems subset highlights the emphasis on advanced textual comprehension in our benchmark. 
The inclusion of Dynamic Prediction subset further ensures comprehensive evaluation under conditions involving temporal changes and challenging visual contexts. 
Collectively, the structured distribution across these subsets fosters a balanced assessment of diverse visual reasoning capabilities crucial for a robust understanding of physical phenomena.

\subsection{Additional Statistics of Dataset}
\begin{figure*}[htbp]
    \centering
    \includegraphics[width=0.85\textwidth]{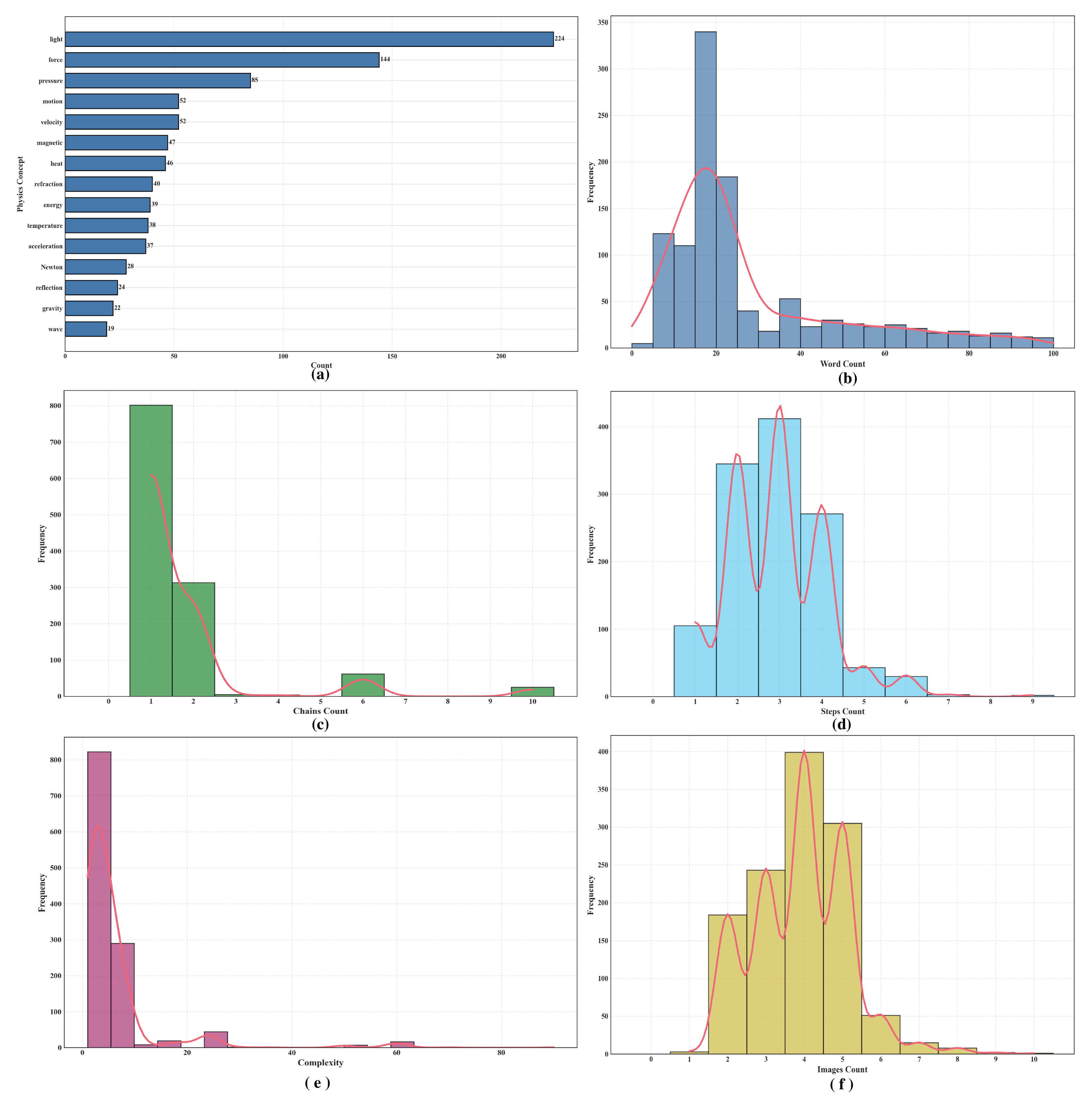}
    \caption{
    \textbf{Additional statistic.}
    Figure a is the Physics Concepts Distribution, this horizontal bar chart shows the frequency of physical concepts that appear in the reasoning steps. The Y-axis represents physical concepts , and the X-axis represents the number of occurrences. 
    Figure b is the Query Word Count Distribution, this histogram shows the distribution of the number of words in the questions.
    The X-axis represents the number of words, and the Y-axis represents the frequency.
    Figure c is the Reasoning Chains Distribution, this histogram shows how many different reasoning paths each sample contains. 
    Figure d is the Reasoning Steps Distribution, this histogram shows how many reasoning steps each sample contains. 
    The X-axis represents the number of steps, and the Y-axis represents the frequency. 
    Figure e is the Reasoning Complexity Distribution, this histogram shows the distribution of complexity indicators.
    Complexity is defined as the number of reasoning steps × the number of different reasoning paths. 
    Figure f is the Sample Images Distribution, this histogram shows how many images each sample contains.
    }
    \label{fig:additional_statistic}
\end{figure*}

This section presents further statistical analyses to offer deeper insights into the composition and characteristics of the dataset. 
As Shown in Figure \ref{fig:additional_statistic}, Figure (a) provides an overview of the distribution of physics concepts encountered within the reasoning steps. 
It reveals that certain foundational concepts such as "light," "force," and "pressure" are notably prevalent, indicating their central importance within the reasoning processes of datasets.
The distribution of these concepts emphasizes their relative significance and highlights the necessity for models to grasp core physics principles robustly.
Figure (b) illustrates the distribution of query word counts through a histogram accompanied by a kernel density estimation curve, effectively capturing the general complexity and length patterns of the queries.
The data suggests a predominance of moderately sized questions, though there exists a notable tail extending towards longer, more complex queries, underscoring the variety in question complexity.

The distribution of reasoning chains, depicted in Figure (c), offers valuable insights into the diversity of dataset in reasoning paths per sample. 
Most samples incorporate one or two distinct chains, highlighting the presence of alternative reasoning pathways. Nonetheless, there is a non-negligible proportion of instances with several reasoning chains, indicating complexity and diversity in the reasoning processes required by the dataset.
Figure (d) examines the distribution of reasoning steps per sample. The analysis indicates variability in the complexity of the reasoning tasks, with most samples containing a moderate number of steps.
This reflects the balance of dataset between simplicity and complexity, essential for comprehensively evaluating reasoning proficiency.

Reasoning complexity, as shown in Figure (e), combines reasoning steps and the number of reasoning chains to provide a composite indicator of overall reasoning demand. 
The distribution confirms that while many instances involve relatively straightforward reasoning, a meaningful subset presents significant complexity, requiring intricate, multi-faceted reasoning capabilities.
Finally, Figure (f) explores the distribution of images included per sample. 
It demonstrates a balanced use of visual information, with most samples featuring several images to guide visual reasoning tasks effectively.
This emphasis on visual context underscores the intent to robustly assess models' capabilities in interpreting and reasoning about visually grounded information.

\section{More Qualitative Examples}
\label{AppendD}
In this section, we present additional qualitative examples to offer deeper insights into our evaluation framework and illustrate the practical application of our metrics.
The examples are organized into three subsections, each highlighting a specific evaluation dimension: Quality, Diversity, and Efficiency. 

\subsection{Examples of Quality Evaluation.}
\begin{figure*}[htbp]
    \centering
    \includegraphics[height=0.7\textheight]{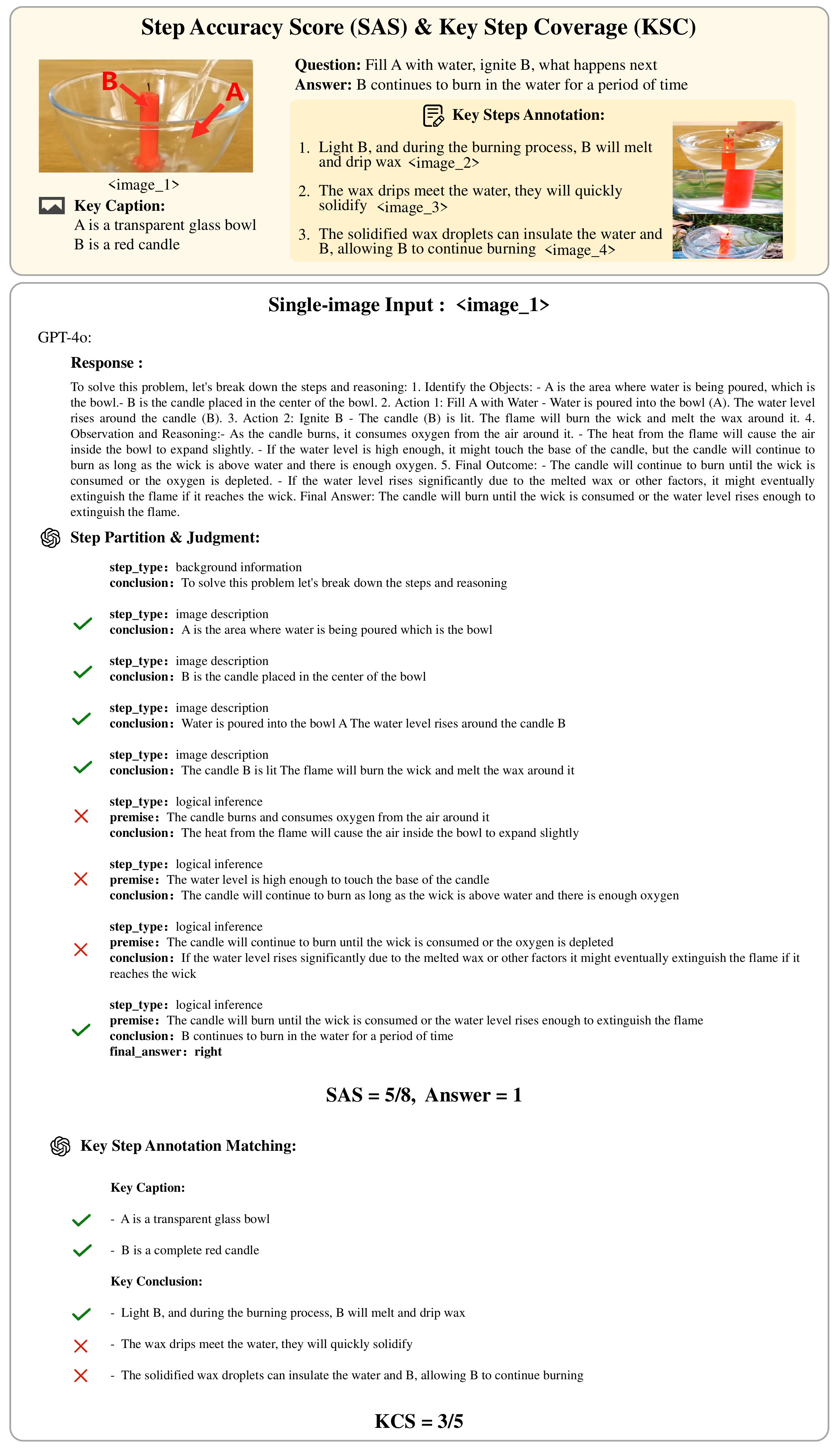}
    \caption{
    \textbf{Examples of Single-Image SAS–KSC Evaluation on Physics Experiment Data.}
    }
    \label{fig:PhyExp_example1}
\end{figure*}

\begin{figure*}[htbp]
    \centering
    \includegraphics[height=0.8\textheight]{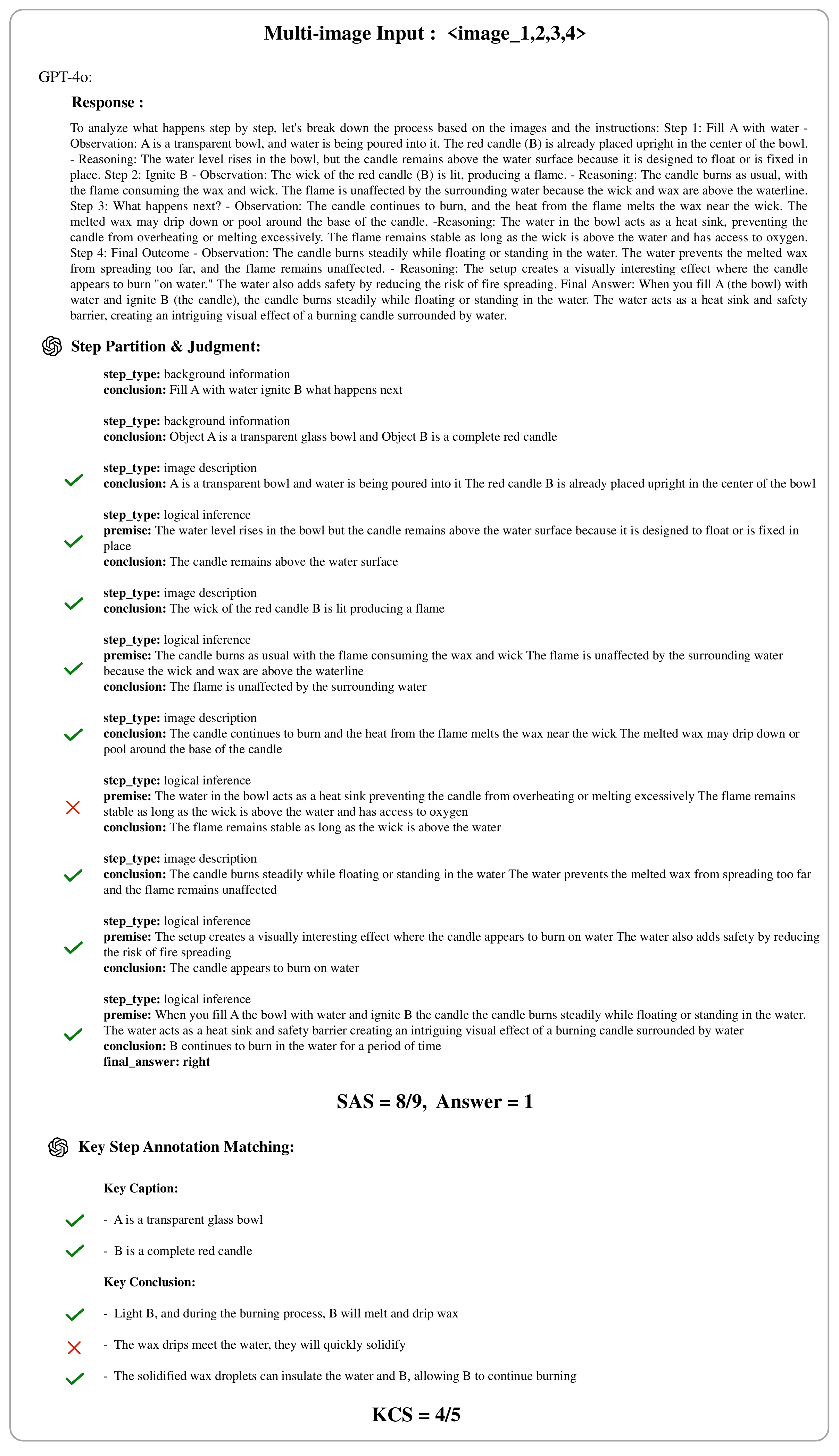}
    \caption{
    \textbf{Examples of Multi-Image SAS–KSC Evaluation on Physics Experiment Data: Enhanced Visual Context Enables Richer Reasoning and Performance Gains.}
    }
    \label{fig:PhyExp_example2}
\end{figure*}

\begin{figure*}[htbp]
    \centering
    \includegraphics[width=\textwidth,height=\textheight, keepaspectratio]{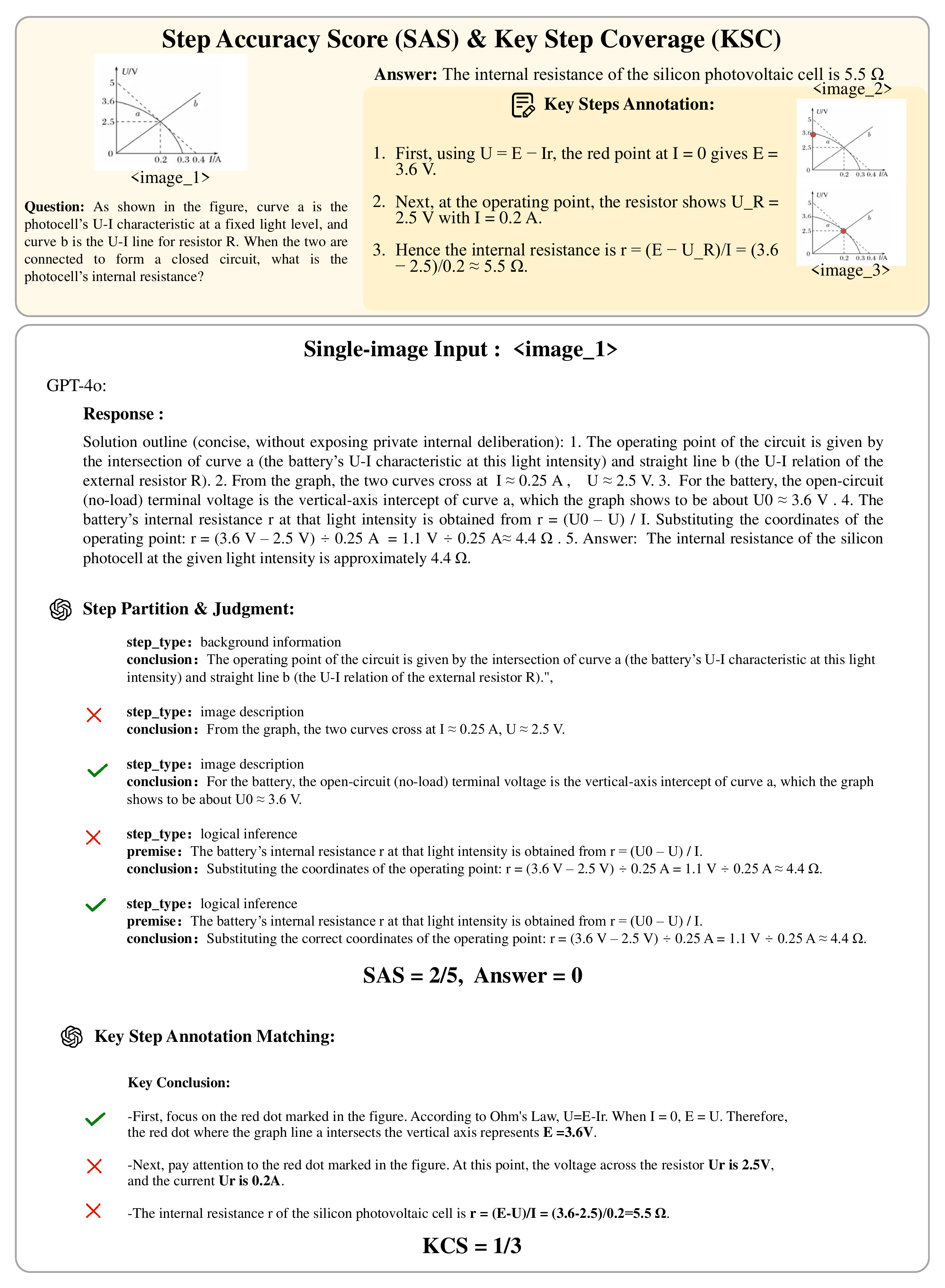}
    \caption{
    \textbf{Examples of Single-Image SAS–KSC Evaluation on Physics Problems Data with GPT-4o.}
    }
    \label{fig:PhyTest1_example1}
\end{figure*}

\begin{figure*}[htbp]
    \centering
    \includegraphics[width=\textwidth,height=\textheight, keepaspectratio]{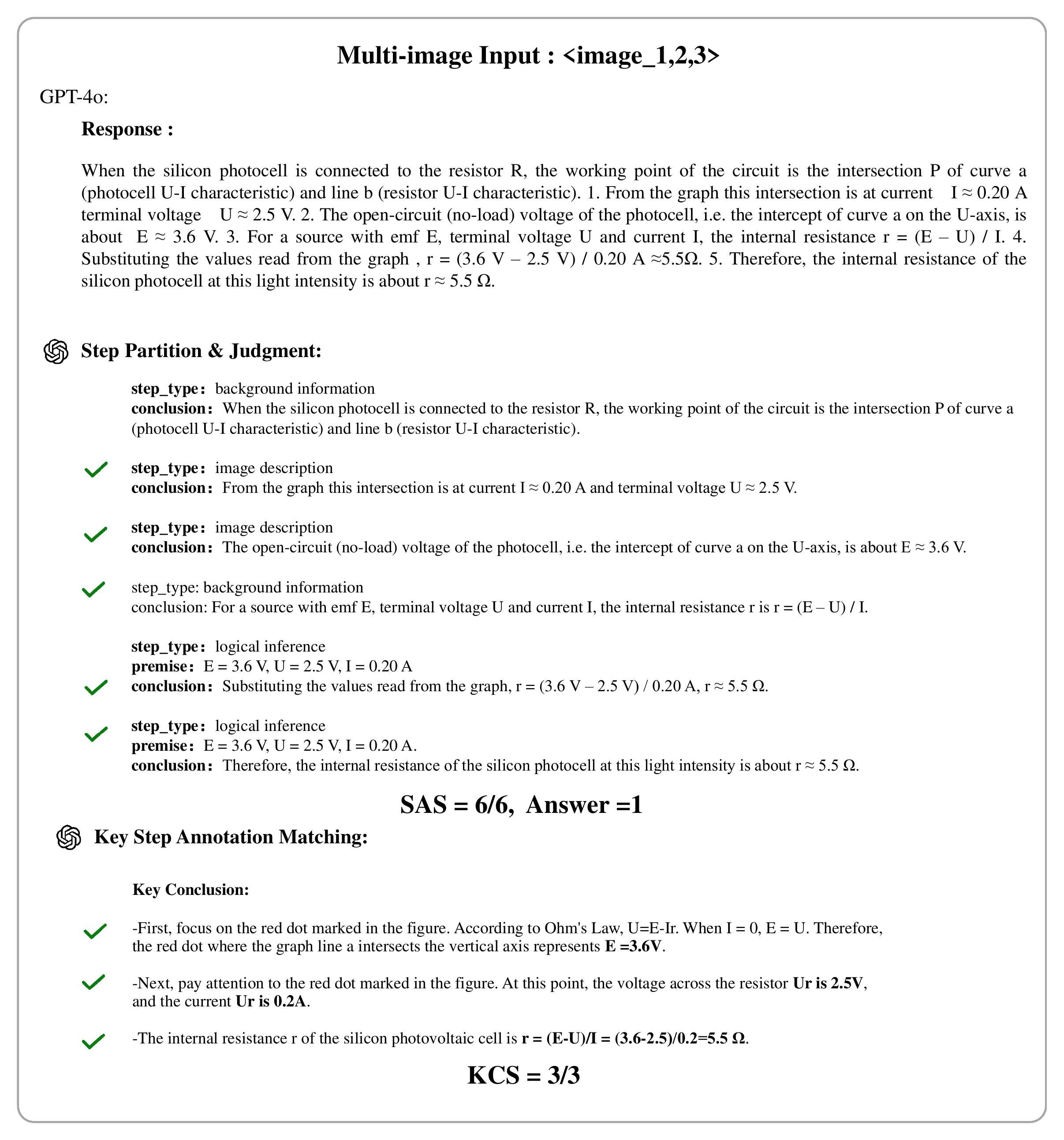}
    \caption{
    \textbf{Examples of Multi-Image SAS–KSC Evaluation on Physics Tasks with GPT-4o: Enhanced Visual Context Boosts Reasoning and Performance.}
    }
    \label{fig:PhyTest1_example2}
\end{figure*}

\begin{figure*}[htbp]
    \centering
    \includegraphics[width=\textwidth,height=\textheight, keepaspectratio]{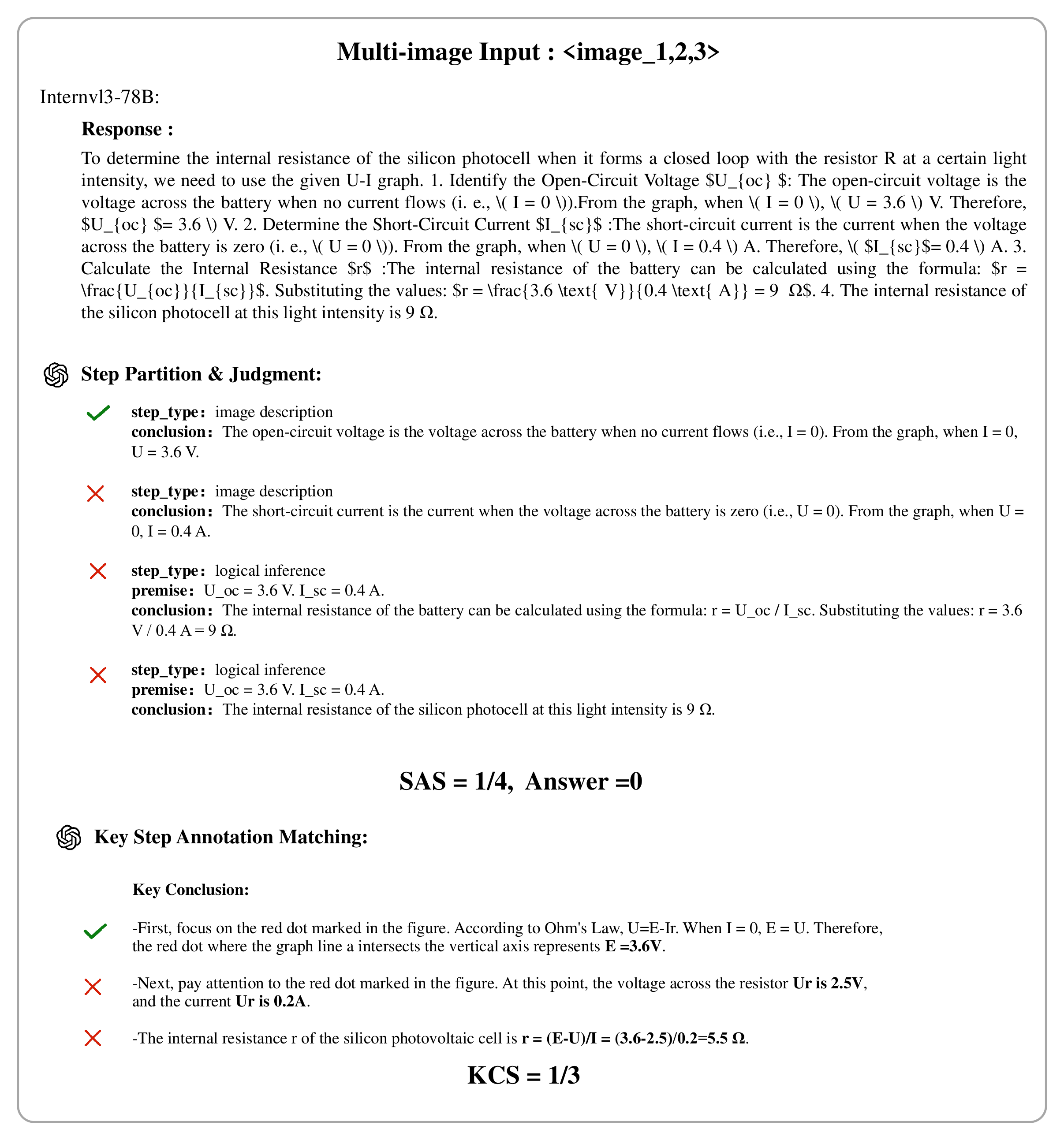}
    \caption{
    \textbf{Examples of Multi-Image SAS–KSC Evaluation on Physics Problems Data with InternVL2.5.}
    }
    \label{fig:PhyTest1_example3}
\end{figure*}

\begin{figure*}[!t]
    \centering
    \includegraphics[width=\textwidth,height=\textheight, keepaspectratio]{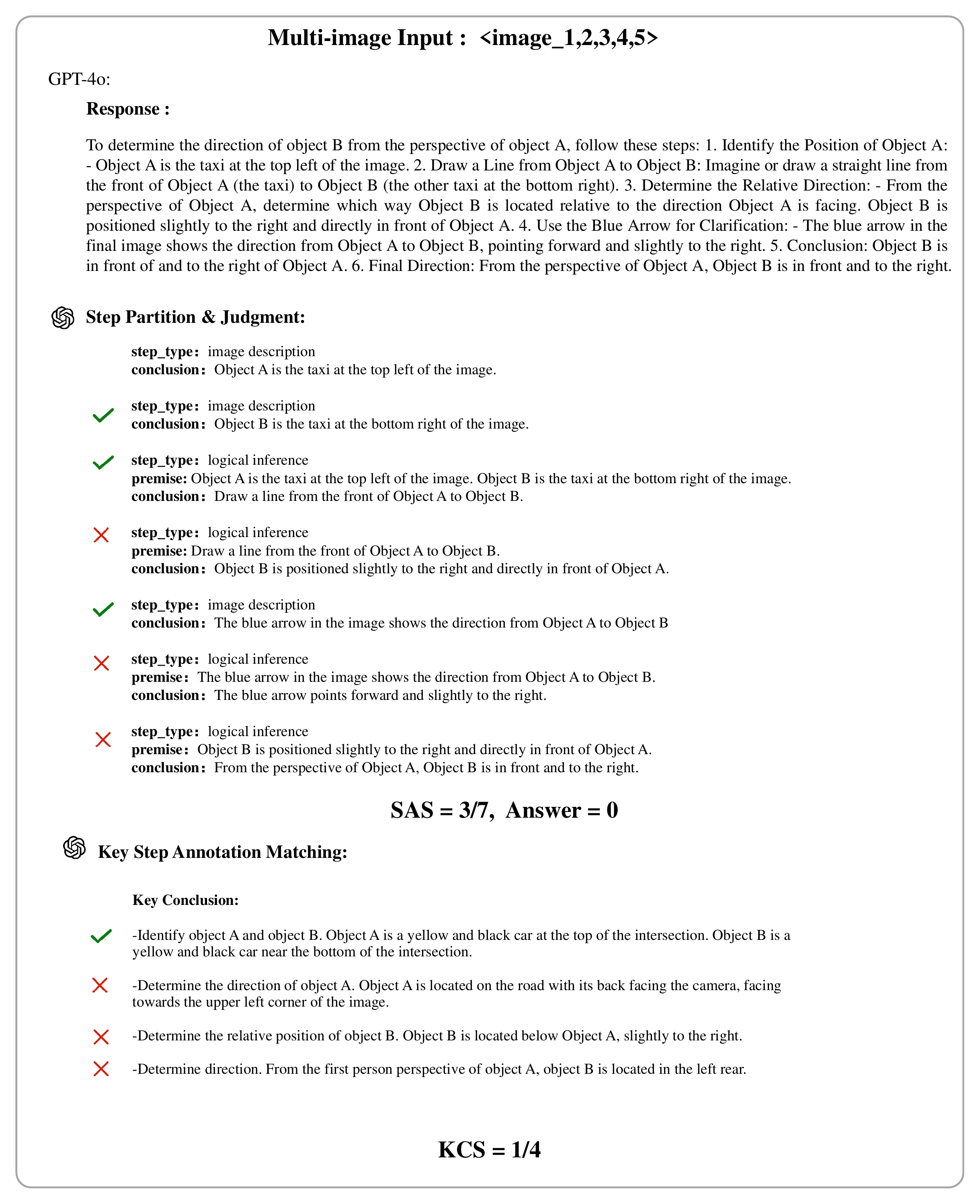}
    \caption{
    \textbf{Examples of Multi-Image SAS–KSC Evaluation on Spatial Relations Data with GPT-4o: Enhenced Visual Context Enables Richer Reasoning and Performance Gains.}
    }
    \label{fig:PhySpatial_example2}
\end{figure*}

\begin{figure*}[!t]
    \centering
    \includegraphics[width=\textwidth,height=\textheight, keepaspectratio]{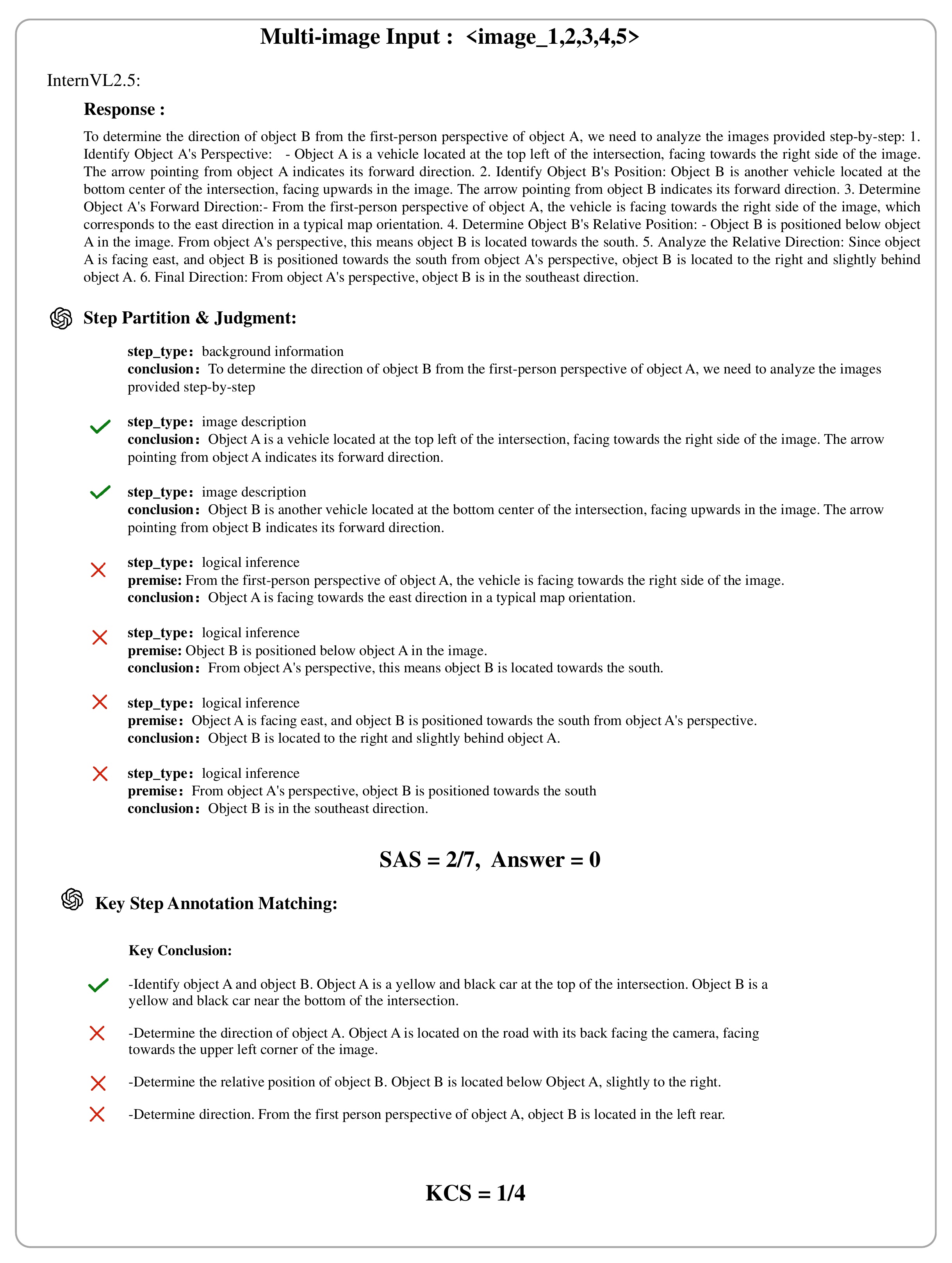}
    \caption{
    \textbf{Examples of Multi-Image SAS–KSC Evaluation on Spatial Relations Data with InternVL3.}
    }
    \label{fig:PhySpatial_example3}
\end{figure*}

This subsection includes eight illustrative examples, as shown in Figure~\ref{fig:PhyExp_example1} to Figure~\ref{fig:PhySpatial_example3}, focusing specifically on evaluating the quality of the reasoning provided by various models.
Each example showcases detailed analyses of model responses, assessing Step Accuracy Score (SAS) and Key Step Coverage (KSC).

\clearpage

\clearpage
\begin{figure*}[!t]
    \centering
    \includegraphics[width=\textwidth,height=\textheight, keepaspectratio]{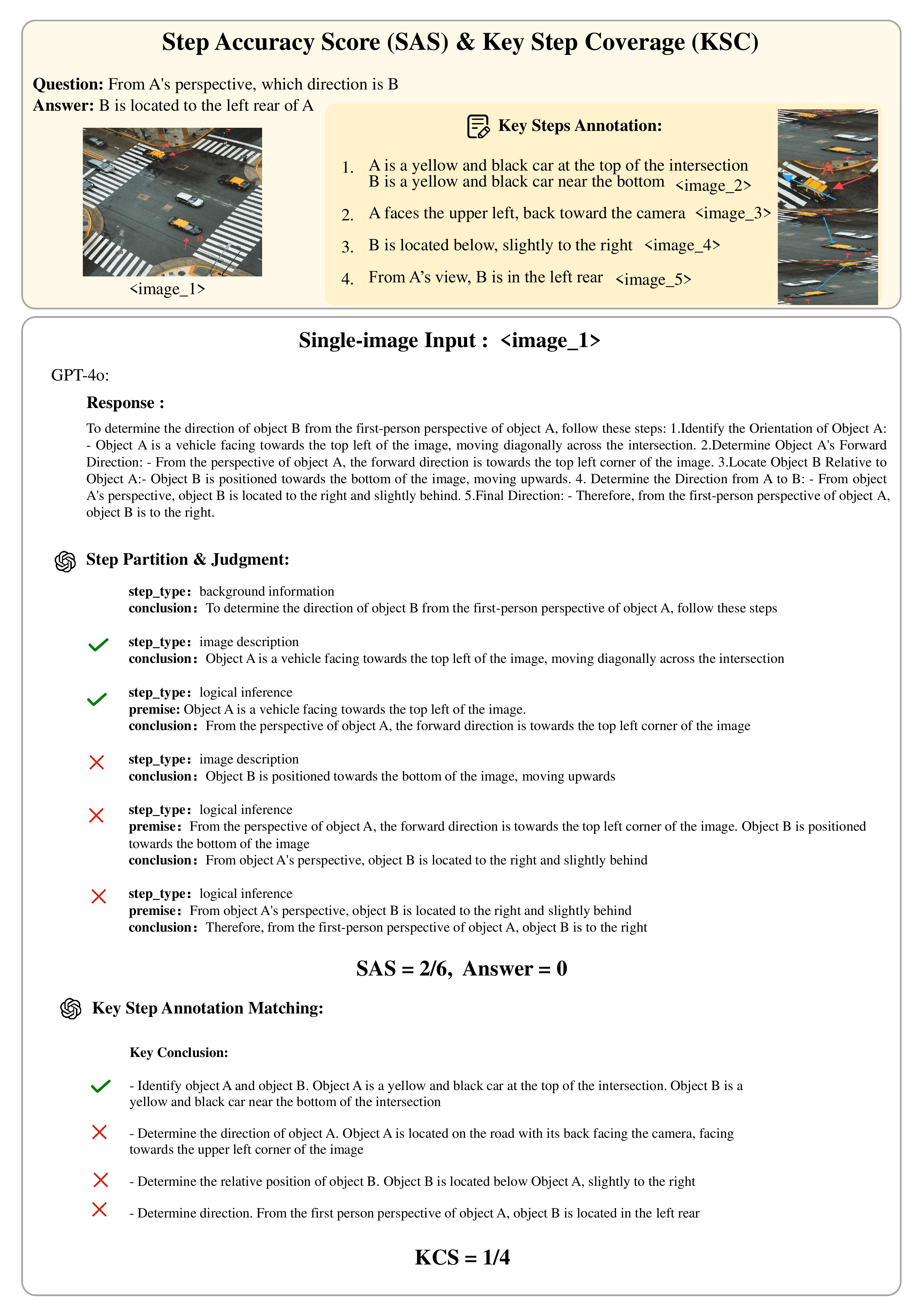}
    \caption{
    \textbf{Examples of Single-Image SAS–KSC Evaluation on Spatial Relations Data with GPT-4o.}
    }
    \label{fig:PhySpatial_example1}
\end{figure*}
\clearpage


\clearpage

\clearpage

\subsection{Examples of Diversity Evaluation.}
\begin{figure*}[htbp]
    \centering
    \includegraphics[width=\linewidth]{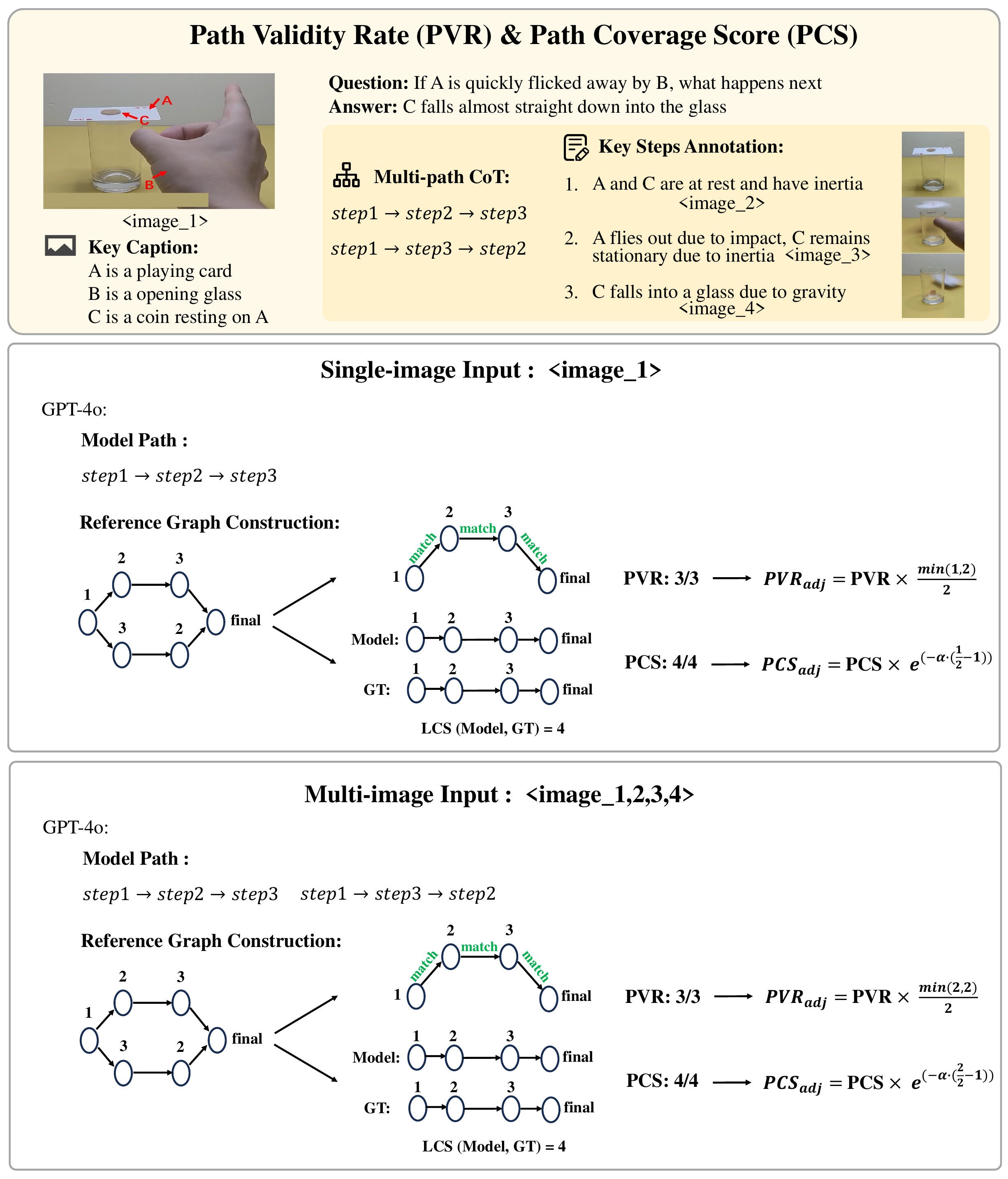}
    \caption{
    \textbf{Examples of Single- and Multi-Image PVR-PCS Evaluation on Spatial Relations Data with GPT-4o.}
    }
    \label{fig:diversity_example}
\end{figure*}

Figure~\ref{fig:diversity_example} provides an example illustrating our diversity evaluation metrics - Path Validity Rate (PVR) and Path Coverage Score (PCS). 
We find that models vary in their ability to explore diverse reasoning paths when processing both single-image and multi-image inputs.

\subsection{Examples of Efficiency Evaluation.}
\begin{figure*}[htbp]
    \centering
    \includegraphics[width=\linewidth]{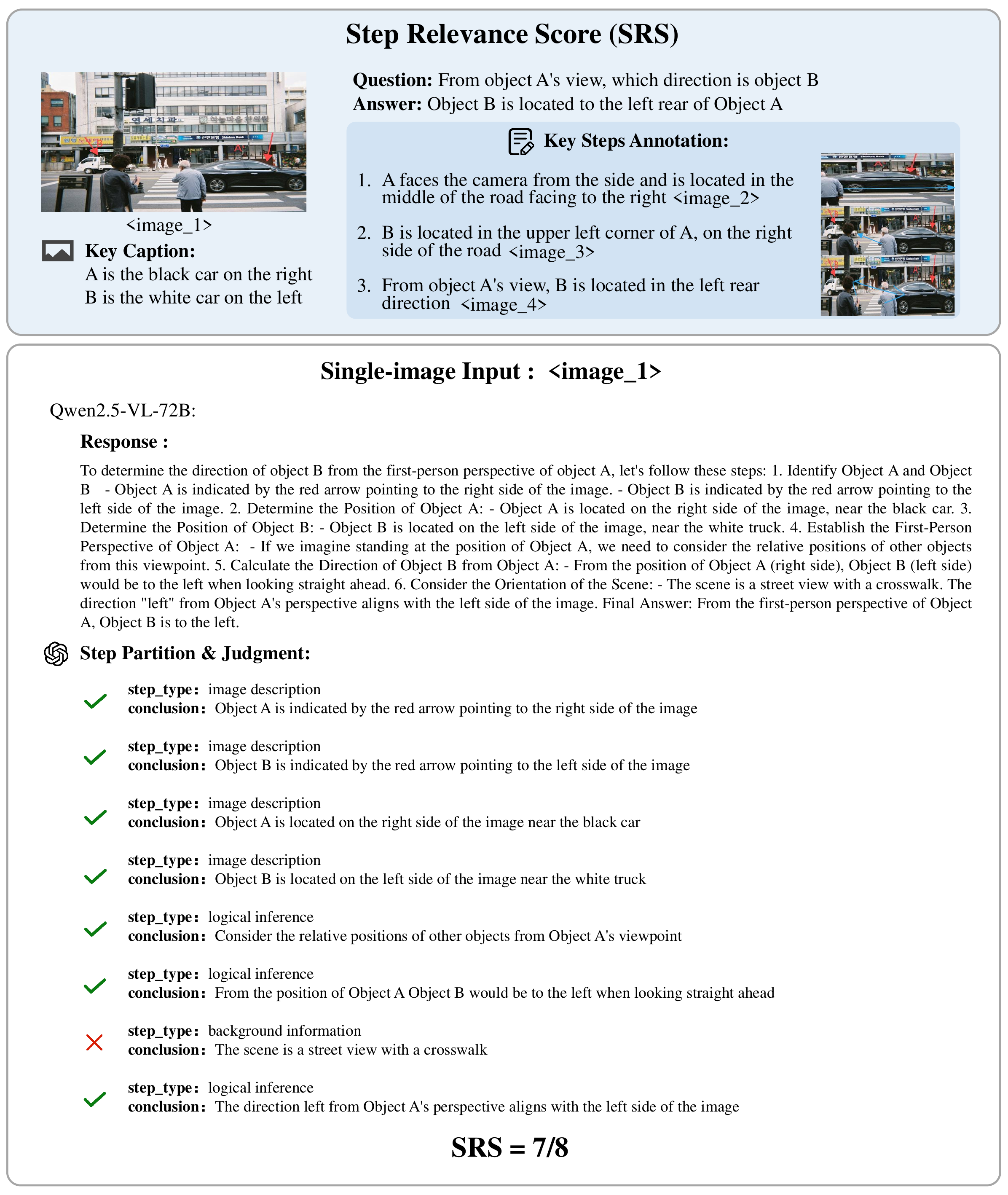}
    \caption{
    \textbf{Examples of Step Relevance Score Evaluation.}
    }
    \label{fig:relevance_example}
\end{figure*}

\begin{figure*}[htbp]
    \centering
    \includegraphics[width=\linewidth]{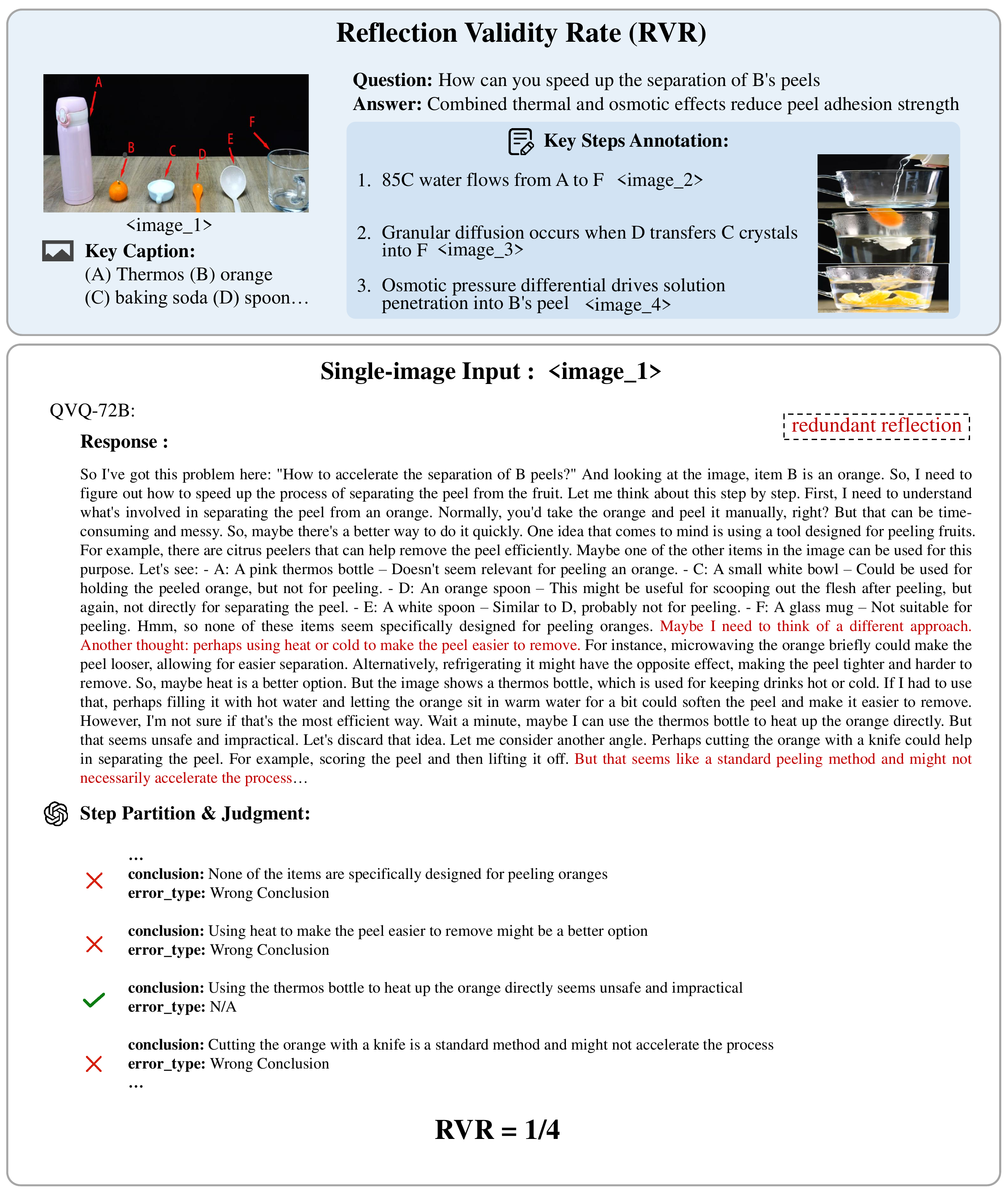}
    \caption{
    \textbf{Examples of Reflection Validity Rate Evaluation.}
    }
    \label{fig:reflection_example}
\end{figure*}

Figure~\ref{fig:relevance_example} 
focuses on evaluating step relevance, clearly indicating how accurately and succinctly the models identify and utilize pertinent information from visual and textual inputs. 
Figure~\ref{fig:reflection_example} specifically illustrates a scenario examining reflection validity, detailing the ability of model to avoid redundant or incorrect reasoning paths.
\clearpage

\section{Limitation}
\label{AppendE}
Our benchmark design, while aiming for rigorous evaluation of visual physical reasoning, inherits several limitations from both dataset construction and evaluation methodology. (i) Scene and Domain Coverage: Despite our efforts to include diverse physical scenarios, MVPBench cannot fully capture the long-tail distribution of real-world physics. This may limit the generalizability of conclusions drawn from our evaluation. To address this, we plan to iteratively expand the dataset with community feedback and new task paradigms. (ii) Annotation Subjectivity: Ground-truth reasoning chains, although carefully curated, may still carry annotator bias in step granularity or interpretation of visual cues. We mitigate this by introducing a graph-based CoT consistency metric to allow flexible yet principled comparisons across models. (iii) Model Usage Constraints: Our evaluation depends on the output of proprietary MLLMs (e.g., GPT-4o), which restricts full control over model internals and fine-tuning procedures. As such, we treat model predictions as black-box outputs and encourage future work to validate findings across both open and closed-source systems for robustness.

\section{Broader impacts}
\label{AppendF}
\textbf{Positive Impacts:}
 On the positive side, this work has the potential to significantly enhance human-AI collaboration in fields such as education, scientific research, and accessibility, by enabling models to perform more transparent and interpretable reasoning across visual and textual modalities.

\textbf{Negative Impacts:}
The potential negative societal impacts of our work are similar to those associated with other MLLMs and LLMs. The development of Visual CoT and MLLMs, while advancing AI, poses societal risks such as increased privacy invasion, the perpetuation of biases, the potential for misinformation, job displacement, and ethical concerns regarding accountability and consent.

\textbf{Mitigation Strategies:}
To mitigate the aforementioned risks, several strategies are considered throughout the development and deployment of our model. First, we adopt a rigorous data curation process aimed at minimizing the propagation of harmful biases, ensuring that training data is as diverse, inclusive, and representative as possible. Second,privacy-preserving techniques such as data anonymization and adherence to data protection regulations (e.g., GDPR) are employed to safeguard user information. Third, we emphasize responsible release practices, including usage guidelines, model cards, and risk documentation, to inform users of the model’s intended scope and limitations. Lastly, we advocate for continued interdisciplinary collaboration with ethicists, legal experts, and affected communities to ensure that the deployment of MLLMs aligns with broader societal values and norms.


\section{Detailed Evaluation prompts}
\label{AppendG}

\subsection{CoT Quality Evaluation Prompts}

\begin{figure*}[!t]
    \centering
    \includegraphics[width=.9\textwidth,height=\textheight, keepaspectratio]{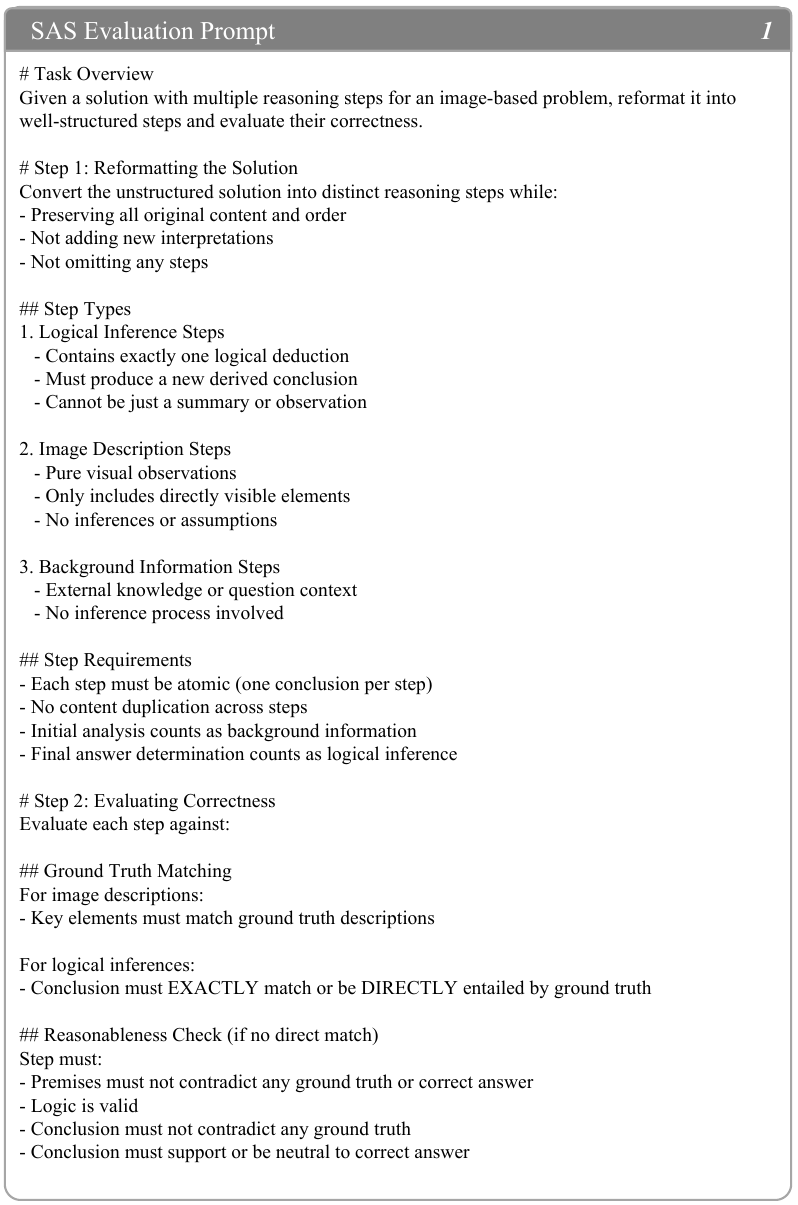}
    \label{fig:PhySpatial_example1}
\end{figure*}
\clearpage

\begin{figure*}[!t]
    \centering
    \includegraphics[width=\textwidth,height=\textheight, keepaspectratio]{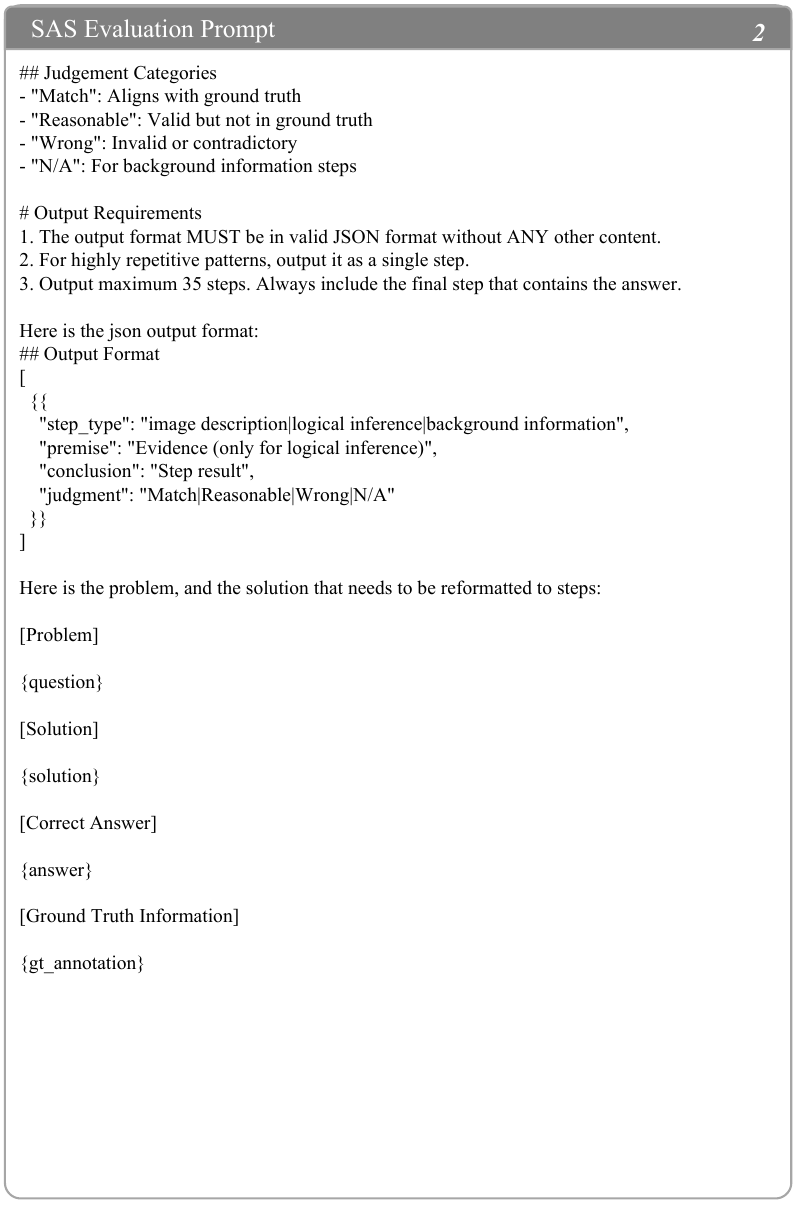}
    \label{fig:PhySpatial_example1}
\end{figure*}
\clearpage

\begin{figure*}[!t]
    \centering
    \includegraphics[width=\textwidth]{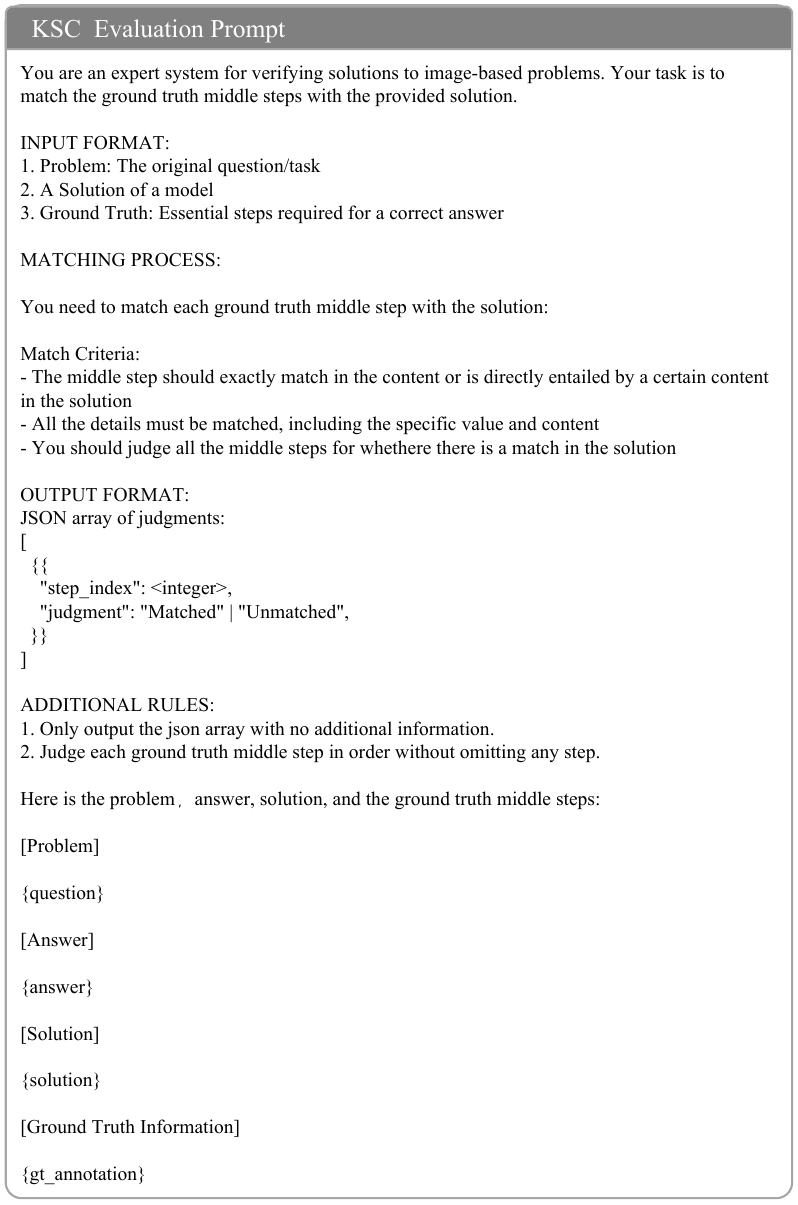}
    \label{fig:PhySpatial_example1}
\end{figure*}

\subsection{CoT Diversity Evaluation Prompts}

\begin{figure*}[!t]
    \centering
    \includegraphics[width=\textwidth,height=\textheight, keepaspectratio]{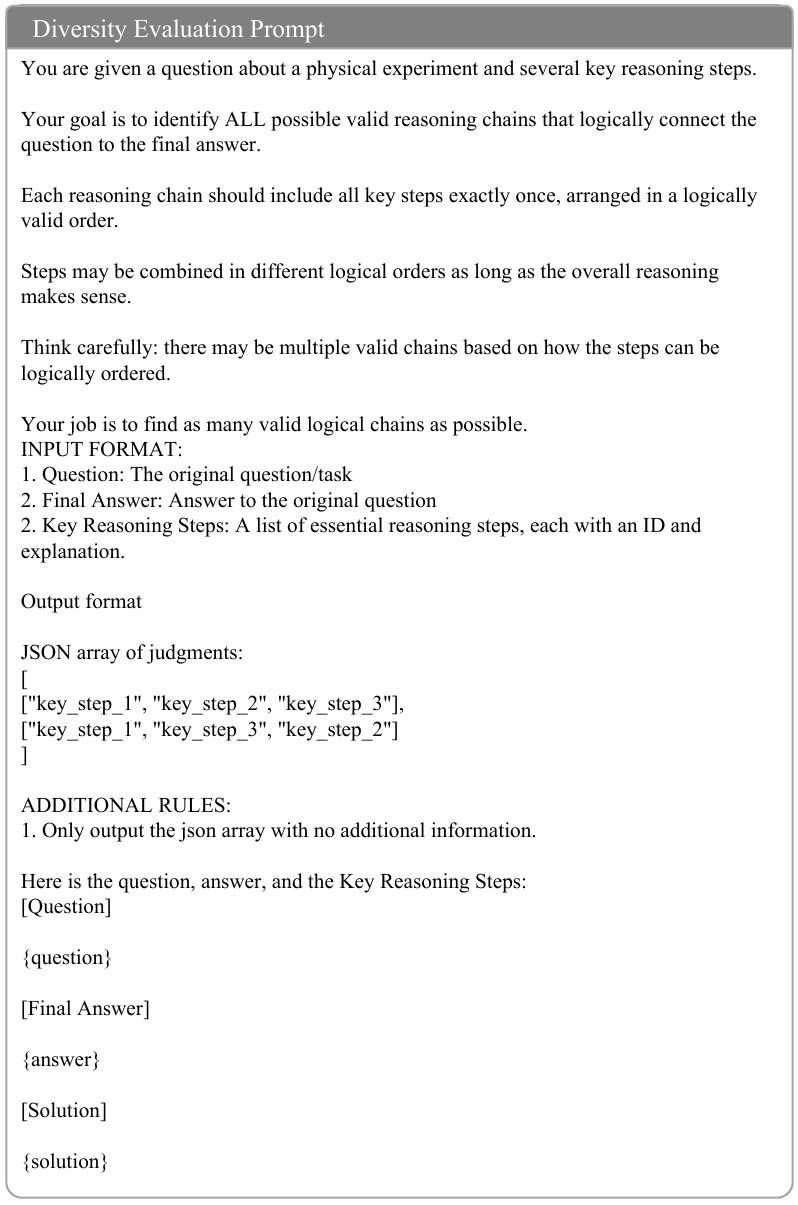}
    \label{fig:PhySpatial_example1}
\end{figure*}
\clearpage

\subsection{CoT Efficiency Evaluation Prompts}
\begin{figure*}[h]
    \centering
    \includegraphics[width=.9\textwidth,height=\textheight, keepaspectratio]{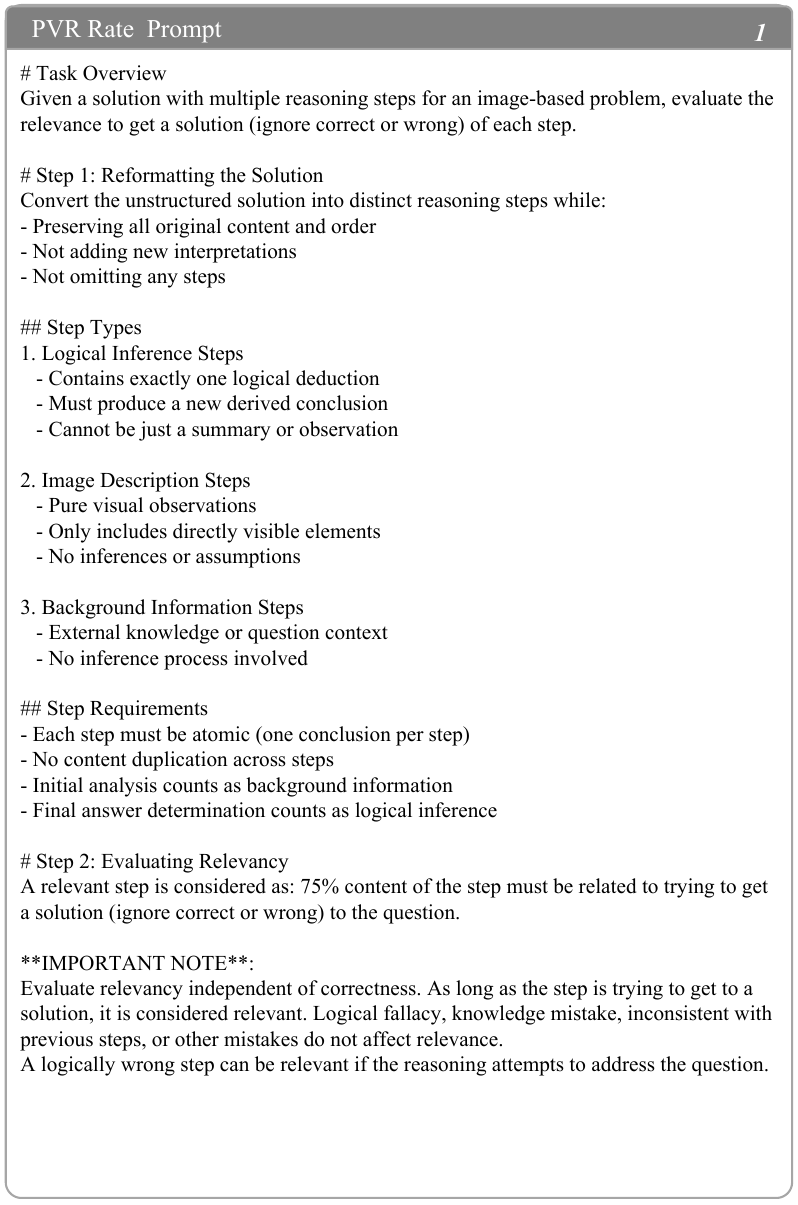}
    \label{fig:PhySpatial_example1}
\end{figure*}
\clearpage

\begin{figure*}[!t]
    \centering
    \includegraphics[width=\textwidth,height=\textheight, keepaspectratio]{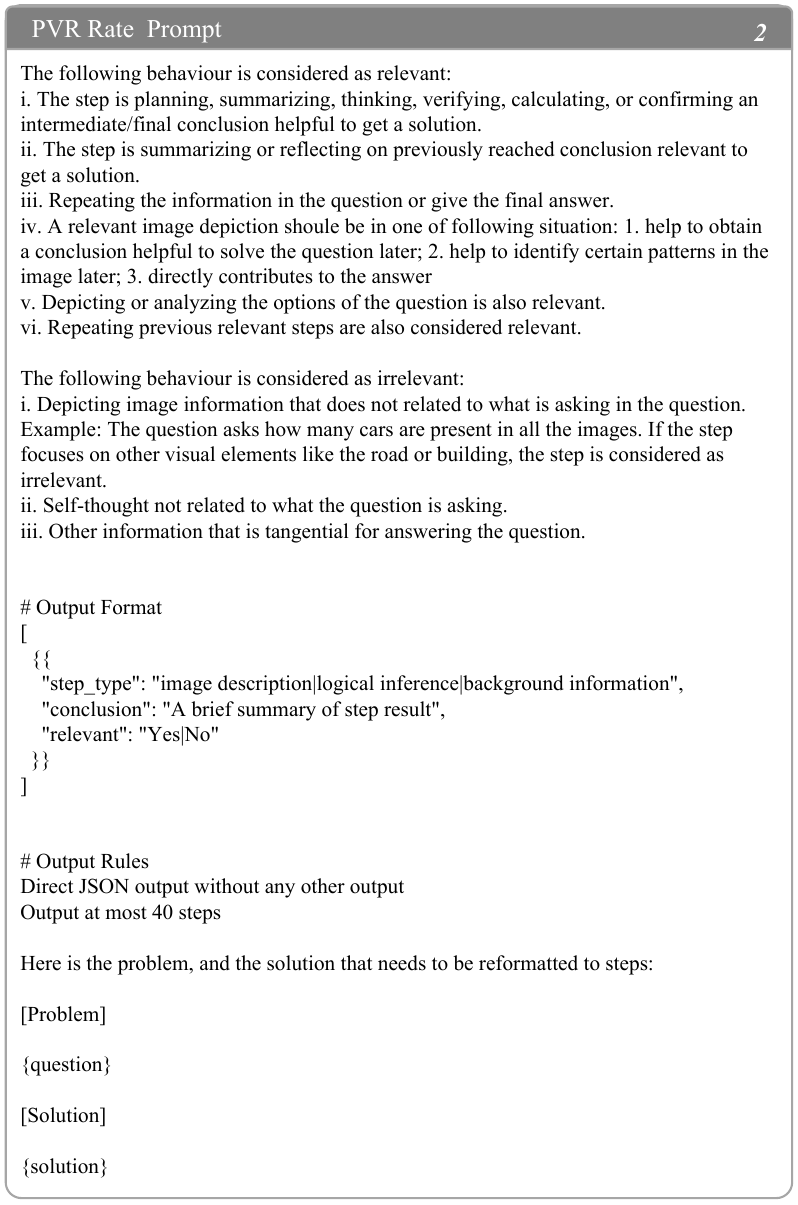}
    \label{fig:PhySpatial_example1}
\end{figure*}
\clearpage

\begin{figure*}[!t]
    \centering
    \includegraphics[width=\textwidth,height=\textheight, keepaspectratio]{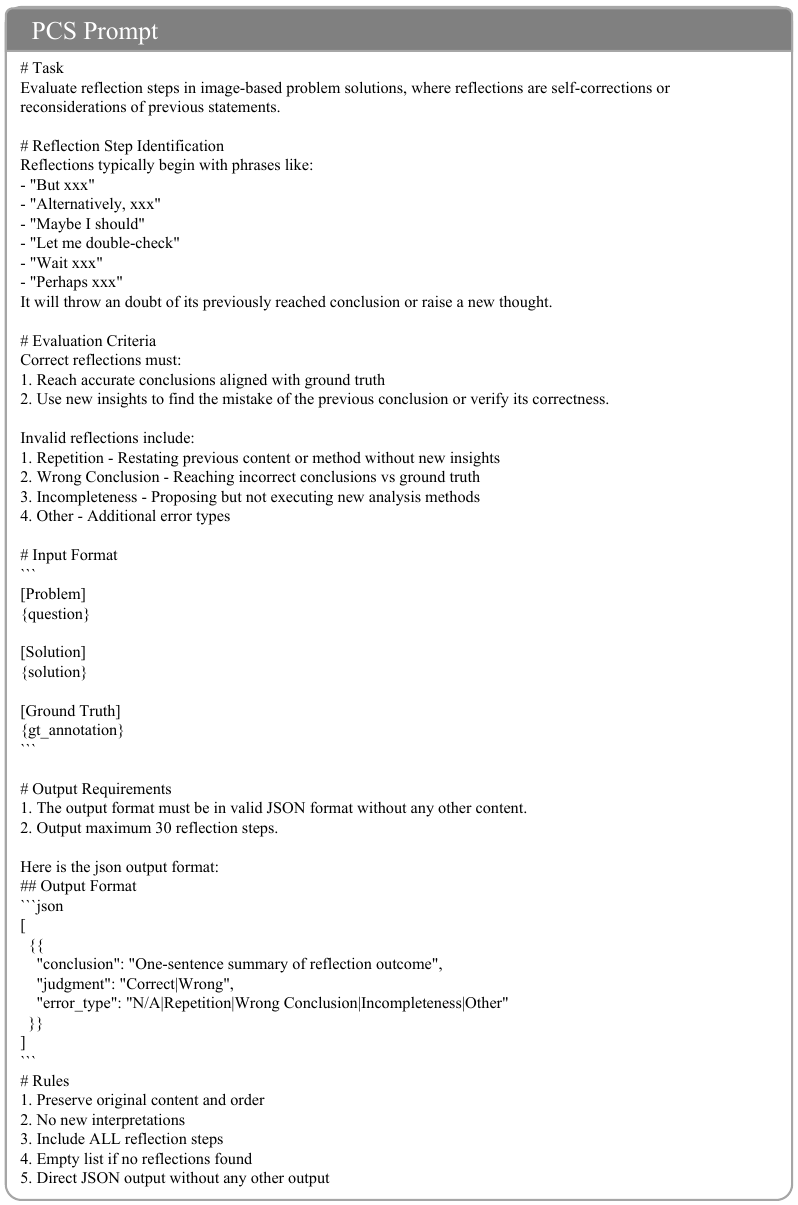}
    \label{fig:PhySpatial_example1}
\end{figure*}
\clearpage

\section{Setup}
\label{AppendH}

\subsection{Experiment Setup}
\label{subsec:exp}

\paragraph{Evaluation Models.}
To comprehensively assess performance on \benchmark, we selected a diverse array of multimodal large language models (MLLMs), encompassing both open-source and closed-source frameworks. 
Among open-source models, we evaluated LLaVA-OV 72B\cite{li2025llavaonevision}, LLaVA-CoT\cite{xu2024llava}, InternVL2.5 78B\cite{chen2024expanding}, InternVL2.5-MPO 78B\cite{wang2024enhancing}, InternVL3 (78B, 78B-Instruct)\cite{InternVL3}, Qwen2.5-VL (7B, 72B)\cite{Qwen2.5-VL}, QVQ-72B\cite{qwen2024qvq}, as well as the recently included Qwen2VL-2B\cite{Qwen2VL}, MM Eureka-7B\cite{mm-eureka}, and R1-VL-2B\cite{r1-vl}, representing various architectures and multimodal integration strategies.
Specifically, InternVL2.5-78B-MPO and InternVL3-78B-Instruct underwent mixed preference optimization (MPO) post-training, while InternVL2.5-78B and InternVL3-78B remained unmodified. 
Furthermore, Qwen2.5VL-7B and Qwen2VL-2B, along with their respective post-trained variants—MM Eureka-7B, which employs large-scale rule-based reinforcement learning (RL), and R1-VL-2B, utilizing Step-wise Group Relative Policy Optimization (StepGRPO)—are of significant interest.
Additionally, prominent closed-source models such as GPT-4o\cite{openai2024gpt4o}, OpenAI o3\cite{openai2025gpto3}, Claude 3.7 Sonnet\cite{claude3.7}, Gemini-2.5\cite{geminiteam2024geminifamilyhighlycapable}, and Grok3\cite{Grok3} were selected based on their state-of-the-art multimodal reasoning capabilities. 
This expanded and carefully curated selection ensures a balanced and thorough evaluation encompassing both openly accessible and proprietary MLLM systems.

\paragraph{Implementation Details.}
All our experiments are conducted under a zero-shot setting, showcasing the generalization capacity of MLLMs for physical reasoning without few-shot prompting or further fine-tuning.
By default, we employ the CoT prompting technique \cite{wei2022chain}, which encourages MLLMs to perform complete reasoning steps for fine-grained evaluation.
All experiments are conducted on NVIDIA V100 GPUs.

\subsection{Model Hyperparameters}
To ensure reproducibility and clarity regarding model settings used during evaluation, Table~\ref{tab:gen_setup} provides detailed information on the hyperparameters and generation setups for each evaluated multimodal large language model (MLLM).
Parameters not explicitly stated indicate that the default settings provided by the respective models were employed. 
This comprehensive specification facilitates transparent comparisons across models and experimental replication.
\begin{table}[htbp]
  \centering
  \caption{\textbf{Generating parameters for MLLMs.} 
    Parameters not explicitly stated indicate the use of the model’s default system settings.}
  \label{tab:gen_setup}
  \begin{tabularx}{\linewidth}{@{}lL@{}}
    \toprule
    \textbf{Model} & \textbf{Generation Setup} \\
    \midrule
    LLaVA-OV-72B
      & \texttt{torch.dtype=torch.float16, max\_new\_tokens=2048, temperature=0.7, device\_map=balanced, min\_pixels=256*28*28, max\_pixels=768*28*28} \\
    LLaVA-CoT
      & \texttt{torch.dtype=torch.float16, max\_new\_tokens=2048, temperature=0.7, device\_map=balanced} \\
    InternVL2.5-78B
      & \texttt{torch.dtype=torch.float16, max\_new\_tokens=2048, temperature=0.7, device\_map=balanced\_low\_0} \\
    InternVL2.5-78B-MPO
      & \texttt{torch.dtype=torch.float16, max\_new\_tokens=1024, temperature=0.7, device\_map=balanced\_low\_0} \\
    InternVL3-78B
      & \texttt{torch.dtype=torch.float16, max\_new\_tokens=1024, temperature=0.7, device\_map=balanced\_low\_0} \\
    InternVL3-78B-Instruct
      & \texttt{torch.dtype=torch.float16, max\_new\_tokens=1024, do\_sample=False, temperature=0.7, device\_map=balanced\_low\_0} \\
    Qwen2.5-VL-7B
      & \texttt{torch.dtype=torch.float16, max\_new\_tokens=1024, do\_sample=False, temperature=0.7, device\_map=balanced} \\
    Qwen2.5-VL-72B
      & \texttt{torch.dtype=torch.bfloat16, temperature=0.7, max\_new\_tokens=1024, device\_map=balanced, min\_pixels=256*28*28, max\_pixels=768*28*28} \\
    QVQ-72B
      & \texttt{torch.dtype=torch.float16, max\_new\_tokens=512, do\_sample=False, temperature=0.7, min\_pixels=256*28*28, max\_pixels=768*28*28, device\_map=balanced, } \\
    MM-Eureka-7B
      & \texttt{torch.dtype=torch.float16, max\_new\_tokens=2048, do\_sample=False, temperature=0.7, device\_map=balanced} \\
    Qwen2VL-2B
      & \texttt{torch.dtype=torch.bfloat16, max\_new\_tokens=2048,  do\_sample=False, temperature=0.7, device\_map=balanced} \\
    R1-VL-2B
      & \texttt{torch.dtype=torch.float16, max\_new\_tokens=2048, use\_cache=True, temperature=0.7} \\
    GPT-4o
      & \texttt{dtype=torch.float16, sampling=False, temperature=0.2, max\_new\_tokens=1024} \\
    OpenAI o3
      & \texttt{dtype=torch.float16, sampling=False, temperature=0.2, max\_new\_tokens=1024} \\
    Claude 3.7 Sonnet
      & \texttt{dtype=torch.float16, sampling=False, temperature=0.2, max\_new\_tokens=1024} \\
    Gemini-2.5-flash-preview-04-17
      & \texttt{dtype=torch.float16, sampling=False, temperature=0.2, max\_new\_tokens=1024} \\
    Grok3
      & \texttt{dtype=torch.float32, sampling=False, temperature=0.2, max\_new\_tokens=1024} \\
    \bottomrule
  \end{tabularx}
\end{table}


\end{document}